\documentclass[11pt,letterpaper]{article}

\usepackage[margin=1in]{geometry}
\usepackage{amsmath,amssymb,amsthm,mathtools}
\usepackage{microtype}
\usepackage[T1]{fontenc}
\usepackage{tikz}
\usepackage{algorithm}
\usepackage{algpseudocode}
\usepackage{float}
\usepackage{tabularx}
\usepackage{booktabs}
\usepackage{amartya_ltx}
\usepackage{aliascnt}
\usepackage[authoryear,round]{natbib}
\definecolor{paperblue}{rgb}{0,0.10,0.45}
\usepackage[
    colorlinks=true,
    linkcolor=paperblue,
    citecolor=paperblue,
    urlcolor=paperblue,
]{hyperref}
\usepackage[capitalize,nameinlink,noabbrev]{cleveref}
\usepackage{comment}

\newcommand{\Lap}{\mathsf{Lap}}
\algnewcommand{\LeftComment}[1]{\Statex  \textbf{#1}\vspace{1mm}}

\newcommand{\DPSAS}{\mathrm{DPSAS}}
\newcommand{\Score}{\mathsf{Score}}
\newcommand{\ExpMech}{\mathrm{ExpMech}}

\newtheorem{theorem}{Theorem}[section]

\theoremstyle{definition}

\newaliascnt{definition}{theorem}
\newtheorem{definition}[definition]{Definition}
\aliascntresetthe{definition}

\newaliascnt{lemma}{theorem}
\newtheorem{lemma}[lemma]{Lemma}
\aliascntresetthe{lemma}

\newaliascnt{proposition}{theorem}
\newtheorem{proposition}[proposition]{Proposition}
\aliascntresetthe{proposition}

\newaliascnt{corollary}{theorem}

\aliascntresetthe{corollary}

\newaliascnt{fact}{theorem}
\newtheorem{fact}[fact]{Fact}
\aliascntresetthe{fact}

\newaliascnt{claim}{theorem}
\newtheorem{claim}[claim]{Claim}
\aliascntresetthe{claim}

\newaliascnt{assumption}{theorem}
\newtheorem{assumption}[assumption]{Assumption}
\aliascntresetthe{assumption}

\newtheorem{remark}{Remark}

\crefname{corollary}{Corollary}{Corollaries}
\Crefname{corollary}{Corollary}{Corollaries}
\crefname{fact}{Fact}{Facts}
\Crefname{fact}{Fact}{Facts}
\makeatletter
\crefname{ALG@line}{Step}{Steps}
\Crefname{ALG@line}{Step}{Steps}
\def\theHALG@line{\thealgorithm.\arabic{ALG@line}}
\makeatother
\crefname{claim}{Claim}{Claims}
\Crefname{claim}{Claim}{Claims}
\crefname{assumption}{Assumption}{Assumptions}
\Crefname{assumption}{Assumption}{Assumptions}
\crefname{remark}{Remark}{Remarks}
\Crefname{remark}{Remark}{Remarks}
\renewenvironment{quote}{%
  \list{}{%
    \rightmargin=0.38cm %
    \leftmargin=0.38cm  %
  }%
  \item\relax
}{%
  \endlist
}

\title{Privacy for Free in Efficient Agnostic Halfspace Learning}

\usepackage{authblk}

\author{Giorgio Racca}
\author{Amartya Sanyal}
\author{Vikrant Singhal}

\affil{Department of Computer Science, University of Copenhagen\\ \texttt{\{g.racca,amsa,vikrant.singhal\}@di.ku.dk}}

\date{\vspace{-3.3em}}

\begin{document}

\maketitle
{\renewcommand{\thefootnote}{}\footnotetext[0]{Authors are listed in alphabetical order.}}

\begin{abstract}
  We provide the first computationally efficient, differentially private algorithms for agnostically learning halfspaces. In particular, we study three different settings of agnostic learning: improper learning under Gaussian and log-concave marginals; proper learning under Gaussian marginals; and semi-agnostic learning under Gaussian marginals. In all three aforementioned settings, the added cost in the sample complexity due to privacy matches the respective non-private sample complexity, implying that we get privacy for free.
\end{abstract}

\section{Introduction}
\label{sec:introduction}
Learning halfspaces is a classical problem in learning theory.
A halfspace is specified by a weight vector \(w\in\reals^d\) and a threshold \(t\in\reals\), and classifies a point \(x\in\reals^d\) according to \(\sgn{\ip{w}{x}+t}\).
We write \(\jointdist\) for the distribution of labeled examples and \(\xmarginal=\jointdist_X\) for its input marginal. In classical \emph{realizable} PAC learning, the distribution \(\jointdist\) is assumed to be linearly separable and the learner is required to output a classifier \(h\) with \(\err[\jointdist]{h}\le \alpha\) for any accuracy parameter \(\alpha\in\br{0,1}\). In the \emph{agnostic} setting, the labels in the dataset may be arbitrary. Thus, the guarantee is relaxed to requiring the learner to output a classifier \(h\) with error \[\err[\jointdist]{h}\le \OPT + \alpha,\] where \(\OPT\) is the minimum error achievable by a halfspace under \(\jointdist\).
This setting allows the data to be noisy or imperfectly separable by a linear decision boundary. Differential privacy (DP)~\citep{dwork2006} imposes an additional constraint, requiring the distribution of the learner's output to change only slightly when a single training example is replaced.
In this paper, we study whether agnostic halfspace learning can be made differentially private with small computational and statistical overhead relative to non-private learning.

Agnostically learning halfspaces is known to be computationally hard, even for non-private improper learners, i.e., learners whose output need not be a halfspace~\citep{tiegel2023hardness}. To overcome this, a rich line of work has focused on distributions with structured marginals but arbitrary labels. Under Gaussian and more generally log-concave marginals,~\citet{kalai2008agnostically}  provided the first algorithm for agnostic learning of halfspaces whose runtime is polynomial in the dimension for every fixed accuracy. For Gaussian marginals, their complexity scales as \(d^{\bigO{1/\alpha^4}}\) which was later improved to \(d^{\bigO{1/\alpha^2}}\) by~\citet{diakonikolas2010bounded}. However, their algorithm was improper. A~\emph{proper} algorithm must output a halfspace and can be desirable because a linear halfspace offers a more compact and interpretable representation of the classifier with \(\bigO{d}\) parameters. For Gaussian marginals,~\citet{diakonikolas2021proper} provided the first agnostic proper learner for halfspaces with somewhat comparable statistical and computational complexity.
For DP agnostic learning, ~\citet{kasivishwanathan2011} provided the
first agnostic private learners for finite classes with sample
complexity scaling logarithmically in the size of the hypothesis
class.~\citet{alon2020} and~\cite{Li2025} provided  general reductions
from agnostic learning to realisable learning. However, their
reductions are not shown to be computationally efficient. For
halfspaces over finite domains,~\citet{beimal2019} provided
efficient private agnostic learners, but over infinite domains such as
\(\reals^d\), distribution-free private learning of halfspaces is
information-theoretically impossible even in the realisable setting
due to their infinite Littlestone
dimension~\citep{bun2015thresholds,alon2019littlestone}.
Distributional assumptions on the marginal \(\xmarginal\) are
therefore essential for both privacy and computational efficiency.
Given the non-private agnostic learners under structured
marginals~\citep{kalai2008agnostically,diakonikolas2021proper}, our
goal in this work is to obtain matching statistical and computational
complexities under differential privacy.

However, even under Gaussian marginals, the \(d^{\bigO{1/\alpha^2}}\) dependence on statistical and therefore computational complexity is likely unavoidable for approaches using polynomial regression due to known SQ lower bounds~\citep{Diakonikolasoptimal2020,GoelSQlowerbound2020}. A natural relaxation for the problem is~\emph{semi-agnostic proper} learning, which allows halfspaces \(h\) with
\[\err[\jointdist]{h}\le C \cdot \OPT + \alpha,\]
where \(C\) is a known universal constant. Non-private learners, under Gaussian marginals, can achieve this guarantee in time polynomial in both \(d\) and \(1/\alpha\) simultaneously~\citep{DiakonikolasKTN22}. This raises the natural question of whether the polynomial dependence on \(d,1/\alpha\) can be maintained for semi-agnostic learning under privacy. Together, these lead to the following question:
\begin{quote}
    \emph{What statistical and computational guarantees can DP algorithms get under Gaussian marginals for i) agnostic improper, ii) agnostic proper, and iii) semi-agnostic proper learning of halfspaces ?}\looseness-1
\end{quote}
\subsection*{Our Results}
\medskip
\noindent\Cref{tab:intro-results} summarizes our results. All algorithms satisfy \(\br{\epsDP,\deltaDP}\)-DP guarantees with respect to the dataset, and their error guarantees hold with high probability. Additionally, all our algorithms run in time polynomial in the sample size for every fixed accuracy.
Throughout, $\softO{\cdot}$ hides factors polylogarithmic in $1/\alpha$,
$1/\beta$, $1/\varepsilon_{\mathrm{DP}}$, and $1/\delta_{\mathrm{DP}}$.

\begin{table}[t]
    \centering
    \small
    \setlength{\tabcolsep}{12pt}
    \begin{tabular}{@{}llll@{}}
        \toprule
        \textbf{Marginal} & \textbf{Learner} & \textbf{Error}          &
        \textbf{Sample complexity $\softO{\cdot}$}                       \\
        \midrule
        Gaussian          & Improper         & $\OPT+\alpha$           &
        $\displaystyle \frac{D_{\mathrm G}}{\alpha^2}
            +\frac{D_{\mathrm G}}{\epsDP\alpha}$                     \\[9pt]
        Log-concave       & Improper         & $\OPT+\alpha$           &
        $\displaystyle \frac{D_{\mathrm L}}{\alpha^2}
            +\frac{D_{\mathrm L}}{\epsDP\alpha^{3/2}}$                   \\[9pt]
        Gaussian          & Proper           & $\OPT+\alpha$           &
        $\displaystyle \frac{D_{\mathrm G}}{\alpha^2}
            +\frac{D_{\mathrm G}}{\epsDP\alpha}
            +\frac1{\alpha^6}
            +\frac1{\epsDP\alpha^5}$                     \\[9pt]
        Gaussian          & Proper           & $C\!\cdot\!\OPT+\alpha$ &
        $\displaystyle \frac d{\alpha^2}
            +\frac{d}
            {\epsDP\alpha}$                                  \\[4pt]
        \bottomrule
    \end{tabular}
    \caption{\small
        The rows correspond, respectively, to
        \Cref{thm:improper-agnostic-gaussian}, \Cref{thm:improper-agnostic-log-concave}, \Cref{thm:private-proper-gaussian}, and \Cref{thm:semi-agnostic}.
        Here, $D_{\mathrm G}\asymp d^{\bigO{1/\alpha^2}}$ and $D_{\mathrm L}\asymp d^{\exp\br{\bigO{1/\alpha}}}$ match known rates in the non-private setting~\citep{diakonikolas2010bounded,kalai2008agnostically}, and $\softOsym$ suppresses polylogarithmic factors.}
    \label{tab:intro-results}
\end{table}

\medskip
\noindent\textbf{Agnostic Improper learning.} Our first result in~\Cref{sec:improper} gives a private improper agnostic learner under the standard Gaussian marginal. The algorithm relies on private polynomial \(\Lone\) regression using clipped Hermite features of degree \(\bigO{1/\alpha^2}\). A key step is to control the effect of clipping on the error and then convert the expected excess-loss bound to a high-probability one. This achieves an error at most \(\OPT + \alpha\) with high probability using sample complexity \begin{equation}\label{eq:improper-gauss-intro}
    n=\softO{\frac{D_p}{\alpha^2} + \frac{D_p}{\alpha\epsDP}}\enskip\text{where}\enskip D_p = \binom{d+p}{p}\end{equation} The first term in~\Cref{eq:improper-gauss-intro} is the statistical cost of the problem due to finite samples and the second term is the cost due to privacy. In~\Cref{app:improper-log-concave}, we extend this approach to known log-concave marginals using \(p=\exp\br{\bigO{1/\alpha}}\), incurring an additional $\alpha^{-1/2}$ factor in the privacy term of~\Cref{eq:improper-gauss-intro} (as shown in~\Cref{tab:intro-results}).

\medskip
\noindent\textbf{Agnostic Proper learning.} In~\Cref{sec:proper-gaussian}, we give a proper DP learner for halfspaces under Gaussian marginals with error at most \(\OPT+\alpha\). As shown in~\Cref{tab:intro-results}, its sample size is \begin{equation}\label{eq:proper-gauss-intro}
    n=\softO{\frac{D_p}{\alpha^2} + \frac{D_p}{\alpha\epsDP} + \frac{1}{\alpha^6} + \frac{1}{\epsDP\alpha^5}}
\end{equation} The learner works by first estimating a polynomial privately, similar to the improper case, which gives the first two terms in~\Cref{eq:proper-gauss-intro}. Then we show that this polynomial can be used to obtain a subspace of dimension \(\bigO{\alpha^{-4}}\) which contains a halfspace with error close to \(\OPT\). Finally, creating a discrete cover of halfspaces supported on this space, and using the exponential mechanism leads to the final output. This subspace creation, discretisation, and exponential mechanism costs the last two terms in~\Cref{eq:proper-gauss-intro}.

\medskip
\noindent\textbf{Semi-Agnostic Proper learning.} In \Cref{sec:semi-agnostic-proper-gaussian}, we give a proper learner under Gaussian marginals with error at most $\bigO{\OPT} + \alpha$. The sample complexity, as described in~\Cref{tab:intro-results}, is
\begin{equation}
    n = \softO{\frac{d}{\alpha^2} + \frac{d}{\alpha\epsDP}}.
\end{equation}
The learner works by utilizing a privatized semi-agnostic learner from~\citet{DiakonikolasKTN22} that provides the error guarantee of $\bigO{\eta}$, where $\eta \geq \OPT$ is known \emph{a priori}. However, since we do not know what $\eta$ is, we search over a small candidate set of $\eta$'s by using the said privatized learner, and then pick the best hypothesis using the exponential mechanism on a fresh validation set.

\begin{remark}
    Note that $d^{O(1/\alpha^2)}\cdot 1/\alpha^c = d^{O(1/\alpha^2)}$. We fix $p$ for both improper and proper agnostic settings, such that it is the smallest $p \in \mathbb{N}$ guaranteeing the existence of the polynomial bases that accurately approximate halfspaces. In other words, we use the same $p$ as required by the known non-private counterparts of our algorithms~\citep{diakonikolas2010bounded,pesenti2026agnostic}. Thus, whatever exponent $d$ has in the sample complexity of the respective non-private learners, we achieve that, while keeping the additional cost in $\poly{1/\alpha}$ explicitly outside the exponent of $d$.
\end{remark}

\begin{remark}
    We additionally note that the added cost in our sample complexity due to DP matches the non-private sample complexity (terms without $\epsDP$ or $\deltaDP$) for each of the three Gaussian settings above when $\epsDP = \Omega(\alpha)$. In that sense, we get privacy \emph{for free}. This is commonly observed in DP statistical estimation (see, e.g.,~\citet{KarwaV18, KamathLSU19}), and such sample complexity is often tight for various problems, such as mean and covariance estimation~\citep{KamathMS22}.
\end{remark}
\noindent\textbf{Notation.}
We denote the class of halfspaces by $\cC$.
For a classifier $h\colon\reals^d\to\bc{\pm1}$, its population
error is $\err[\jointdist]{h}\coloneqq
    \bP_{(X,Y)\sim\jointdist}[h(X)\neq Y]$ and, given a sample $S$, its empirical error is $\herr[S]{h}\coloneqq \abs{S}^{-1}\sum_{(x,y)\in S}\mathbf{1}\{h(x)\neq y\}$.
Unless stated otherwise, the input marginal is $\mu = \mathcal{N}(0, I_d)$; the only
exception is our result for log-concave marginals (\Cref{thm:improper-agnostic-log-concave}).
We define $\OPT\coloneqq \inf_{c\in\cC}\err[\jointdist]{c}$.
For a function $g\colon\reals^d\to\reals$ and
$q\in\bc{1,2}$, we write
$\Lpnorm[\xmarginal]{q}{g}\coloneqq
    (\bE_{X\sim\xmarginal}[\abs{g(X)}^q])^{1/q}$.
The formal definition of differential privacy and associated properties are given in~\Cref{subsec:privacy-tools}.

\section{Private Improper Agnostic Learning}
\label{sec:improper}
In this section, we give a differentially private algorithm for agnostically learning halfspaces under Gaussian marginals.
We build on the polynomial $\Lone$ regression algorithm of \citet{kalai2008agnostically}, which fits a low-degree polynomial to the labels and then thresholds its predictions.

To privatize this approach, we need two additional properties. We
first need a bounded domain that contains the coefficients of a good
polynomial approximation to every halfspace. Working in the
orthonormal Hermite basis, the unit-norm approximation guarantee of
\Cref{lem:unit-norm-approx-gauss}
allows us to restrict the coefficient vector to the Euclidean unit
ball.  Second, we need uniformly bounded feature norms so that the
regression loss is Lipschitz in the coefficient vector. We obtain this
by clipping the feature vectors. This increases the population $\Lone$
loss by at most the probability of clipping, which we bound
in~\Cref{lem:hermite-clipping}. These two properties allow us to fit
the coefficients using private convex optimisation~\citep{kulkarni2021private}. We then
repeat the DP optimisation step on independent samples to boost its
success probability and use the exponential mechanism~\citep{mcsherry2007mechanism} to privately select
among the resulting thresholded classifiers. Our main result is stated as follows.

\begin{theorem}[DP improper agnostic learner for Gaussian marginals]
  \label{thm:improper-agnostic-gaussian}
  Let \(\alpha,\beta,\epsDP,\deltaDP\in(0,1/2]\).  Set
  \(p=\lceil C\alpha^{-2}\rceil\) for a sufficiently large universal
  constant \(C>0\), and \(D_p=\binom{d+p}{p}\).  Suppose \(d\ge
  C_{\mathrm{dim}}\alpha^{-5}\), where \(C_{\mathrm{dim}}>0\) is a
  sufficiently large universal constant.  Let \(P\) be any
  distribution on \(\reals^d\times\bc{\pm1}\) with marginal
  \(\mu=\mathcal N(0,I_d)\).  Then
  \Cref{alg:private-polynomial-regression} is
  $(\epsDP,\deltaDP)$-DP. Given
  \[
    n=\softO{
      \frac{D_p}{\alpha^2}
      +\frac{D_p}{\epsDP\alpha}
    }
  \]
  i.i.d.\ samples from $\jointdist$, it returns a classifier $h$ satisfying
  \(
  \err[\jointdist]{h}
  \le \OPT+\alpha
  \)
  with probability at least $1-\beta$.
  The algorithm runs in time $\operatorname{poly}(n)$.
\end{theorem}

\noindent The two terms in the sample complexity arise from
statistical estimation and private optimization, respectively.  When
$\epsDP\ge\alpha$, the privacy term is absorbed into the statistical
term, so privacy is free up to logarithmic factors.  The sample
complexity $\softO{D_p/\alpha^2}$ matches the dependence on $d$ of
known algorithms for non-private agnostic learning of halfspaces under
Gaussian marginals~\citep{diakonikolas2010bounded}.  Moreover, the
$d^{{1/\alpha^2}}$ cost is necessary in the statistical query
model, even without privacy~\citep{Diakonikolasoptimal2020}.
In~\Cref{app:improper-log-concave}, we extend the method to
log-concave marginals and establish an analogous guarantee.

\subsection{The Algorithm}
\label{sec:improper-algorithm}

We represent each input $x$ by its feature vector
$\psi_p(x)=(h_\kappa(x))_{|\kappa|\le p}\in\mathbb R^{D_p}$ of
normalized Hermite polynomials~\citep{ODonnell_2014} of degree at most
\(p\). The degree $p=\lceil C\alpha^{-2}\rceil$ suffices to
approximate every halfspace in $\Lone[\xmarginal]$ up to error
$\alpha$~\citep{diakonikolas2010bounded}.  To bound the influence of
any single example and make the regression loss Lipschitz in its
coefficients, we clip the feature vectors to norm at most
$T=2\sqrt{D_p}$:
\(
\bar\psi_{p,T}(x)
\coloneqq
T\psi_p(x)/\max\{T,\|\psi_p(x)\|_2\}.
\) Then, by \Cref{lem:hermite-clipping}, the clipping
probability is at most $\bigO{\alpha}$. Given a sample $S$, the regression step approximately minimizes
the empirical $\Lone$ loss
\begin{equation}\label{eq:clipped-regre}
  \widehat L_S(v)
  =\frac{1}{|S|}\sum_{(X_i,Y_i)\in S}
  \bigl|\langle v,\bar\psi_{p,T}(X_i)\rangle-Y_i\bigr|
\end{equation}
over the unit ball $B_2^{D_p}=\{v\in\mathbb R^{D_p}:\|v\|_2\le1\}$.
This loss is convex and $T$-Lipschitz in $v$, so we apply the private
convex ERM algorithm of \citet{kulkarni2021private}, restated in
\Cref{lem:private-lipschitz-erm}.  A potential alternative approach is
to implement polynomial regression using statistical
queries~\citep{kearns1998efficient} and then simulate these queries
privately~\citep{kasivishwanathan2011}. For the clipped regression
problem of~\Cref{eq:clipped-regre}, these queries can estimate the
expected subgradients. However, making this SQ implementation explicit
and replacing its query answers with actual private estimates leads to
an algorithm with basically the same structure as ours. For this
reason, we do not write the SQ version of this algorithm here.

Next, the regression guarantee holds in expectation, so we repeat the
regression step on independent training blocks to ensure that at least
one output has small population loss with high probability.
Specifically, \Cref{alg:private-polynomial-regression} first partitions
the sample into $k$ training blocks of size $n_0$ and a validation
block of size $m$.  It runs the private optimizer on each training
block to obtain coefficient vectors $\hat v_1,\dots,\hat v_k$ with
scores $\ip{\hat v_j}{\bar\psi_{p,T}(x)}$, and then thresholds them at
$M$ evenly spaced points in $[-1,1]$ to produce $kM$ candidate
classifiers.  Finally, it selects one candidate using the exponential
mechanism~\citep{mcsherry2007mechanism}, scoring each candidate by its
fraction $\herr[S_{\mathrm{val}}]{\cdot}$ of misclassified validation
examples.  Repeating the regression step on $k$ blocks boosts its
success probability, and the validation step chooses the coefficient
vector and the threshold jointly.

\begin{algorithm}[t]
  \caption{Private polynomial $\Lone$ regression}
  \label{alg:private-polynomial-regression}
  \begin{algorithmic}[1]
    \Require Sample $S$ of size $n$; parameters $\alpha,\beta,\epsDP,\deltaDP$.
    \State Set $p=\lceil C\alpha^{-2}\rceil$, $T=2\sqrt{D_p}$,
    $k=\lceil\log_2(2/\beta)\rceil$, $M=\lceil C/\alpha\rceil$, and $n_0,m$ as in
    \eqref{eq:gaussian-training-size}--\eqref{eq:gaussian-validation-size}.
    \State Partition $S$ into training blocks $S_1,\dots,S_k$ of size $n_0$ and a validation block
    $S_{\mathrm{val}}$ of size $m$.
    \For{$j=1,\dots,k$}
    \State $\hat v_j\gets$ output of the $(\epsDP,\deltaDP)$-DP
    optimizer on $\widehat L_{S_j}$ over $B_2^{D_p}$.
    \EndFor
    \State For $(j,\ell)\in[k]\times[M]$, set $t_\ell=-1+(2\ell-1)/M$ and
    $h_{j,\ell}(x)=\operatorname{sign}\bigl(\langle\hat v_j,\bar\psi_{p,T}(x)\rangle-t_\ell\bigr)$.
    \State Sample $(\hat\jmath,\hat\ell)$ with probability proportional to
    $\exp\bigl(-\epsDP\, m\,\herr[S_{\mathrm{val}}]{h_{j,\ell}}/2\bigr)$.
    \State \Return $h_{\hat\jmath,\hat\ell}$.
  \end{algorithmic}
\end{algorithm}

\subsection{Analysis}
\label{sec:improper-accuracy}

In this section, we outline the accuracy analysis; the full accuracy and privacy proofs are deferred
to~\Cref{app:improper-gaussian}.  Define the population loss
\(
L_\jointdist(v)\coloneqq\mathbb E_{(X,Y)\sim \jointdist}\bigl|\langle v,\bar\psi_{p,T}(X)\rangle-Y\bigr|
\) and its minimum over the unit ball
\(
L^\star\coloneqq\min_{\|v\|_2\le1}L_\jointdist(v).
\) For a universal constant \(c\in(0,1/4]\), we first compare the minimum regression loss with the error of the best halfspace:
\begin{equation}
  L^\star              \le 2\,\OPT+2c\alpha. \label{eq:Lstar-bound}
\end{equation}
We then show that the algorithm returns a classifier whose error satisfies
\begin{equation}
  \err[\jointdist]{h}  \le \tfrac12 L^\star+3c\alpha\label{eq:err-Lstar}
\end{equation} with probability at least \(1-\beta\). Together, these give $\err[\jointdist]{h}\le\OPT+4c\alpha\le\OPT+\alpha$ with the same probability. \looseness=-1%

We begin with \Cref{eq:Lstar-bound}, which compares the clipped regression problem with the best halfspace.
Fix $f\in\mathcal C$.
By~\Cref{lem:unit-norm-approx-gauss}~\citep{diakonikolas2010bounded,pesenti2026agnostic}, there is a polynomial $q=\langle v_q,\psi_p\rangle$ with $\|v_q\|_2\le1$ and $\Lonenorm[\xmarginal]{q-f}\le c\alpha$.
Let $s(x)\coloneqq T/\max\{T,\|\psi_p(x)\|_2\}$, so that $\bar\psi_{p,T}(x)=s(x)\psi_p(x)$.
Since
$\langle v_q,\bar\psi_{p,T}(x)\rangle=s(x)q(x)$ with $0<s(x)\le1$, and
$|y|=1$,
\[
  \begin{aligned}
    |s(x)q(x)-y|
     & \le s(x)|q(x)-y|+1-s(x)          \\
     & \le |q(x)-f(x)|+|f(x)-y|+1-s(x).
  \end{aligned}
\]
The first inequality follows by adding and subtracting \(s(x)y\) and applying the triangle inequality.
Taking expectations gives
\[
  L_\jointdist(v_q)\le 2\err[\jointdist]{f}+\Lonenorm[\xmarginal]{q-f}+\mathbb E_\xmarginal[1-s(X)].
\]
The three terms are the classification error of $f$, the polynomial
approximation error, and the clipping bias.  Since $1-s(x)\le\mathbf
  1\{\|\psi_p(x)\|_2>T\}$, \Cref{lem:hermite-clipping} bounds the last
term by $c\alpha$.  Clipping thus enters only through its probability.
The approximation error is also at most $c\alpha$, so
$L_\jointdist(v_q)\le 2\err[\jointdist]{f}+2c\alpha$.  As
$\|v_q\|_2\le1$, we have $L^\star\le L_\jointdist(v_q)$, and taking
the infimum over $f$ gives \eqref{eq:Lstar-bound}.\looseness-1

For \Cref{eq:err-Lstar}, we first show that at least one of the $k$
regression runs is nearly optimal. The loss is convex and
$T$-Lipschitz and thus for each training block,
\Cref{lem:private-lipschitz-erm} from~\citet{kulkarni2021private}
bounds the expected optimization error. Then, standard Rademacher complexity bound for linear predictors under $L_1$ loss~(\Cref{lem:expected-regression-deviation}), applied with $\mathbb
  E_\xmarginal\|\bar\psi_{p,T}(X)\|_2^2\le D_p$, bounds the expected
deviation between the empirical and population losses. We choose $n_0$
so that these two contributions sum to at most $c\alpha$ (see
\eqref{eq:gaussian-training-size}), giving
\[
  \mathbb E\bigl[L_\jointdist(\hat v_j)\bigr]-L^\star\le c\alpha,
\]
where the expectation is over $S_j$ and the optimizer's randomness.
Since $\hat v_j$ lies in the
unit ball, its excess loss $L_\jointdist(\hat v_j)-L^\star$ is
non-negative. Markov's inequality therefore gives
$L_\jointdist(\hat v_j)\le L^\star+2c\alpha$ with probability at least $1/2$. The $k$ runs are independent,
so with probability at least $1-2^{-k}\ge1-\beta/2$, some run $j^\star$ satisfies this bound.

We next show that thresholding this run yields a good
candidate classifier. For any fixed regression output $\hat v_j$, a uniformly
random threshold $\tau\in[-1,1]$ gives expected classification error
at most $L_\jointdist(\hat v_j)/2$, where the expectation is over
$\tau$~\citep[see also \Cref{lem:random-threshold}]
{kalai2008agnostically}. Rounding $\tau$ to the midpoint of its grid
interval makes it uniform over the $M$ grid thresholds and increases
the expected classification error by at most $1/(2M)\le c\alpha$. Thus, at least one grid threshold has error no larger than this
average. On the event that some run $j^\star$ satisfies this loss
bound, we have that\[
  \min_{j,\ell}\err[\jointdist]{h_{j,\ell}}
  \le\tfrac12L_\jointdist(\widehat v_{j^\star})+c\alpha
  \le\tfrac12L^\star+2c\alpha.
\]

This establishes that the candidate set contains a good classifier. It
remains to select one privately. The algorithm uses the exponential
mechanism on the independent validation sample $S_{\mathrm{val}}$ to
select among the $kM$ candidate classifiers. By
\Cref{lem:private-finite-validation}, with $m$ as in
\eqref{eq:gaussian-validation-size}, the selected classifier has error
at most $c\alpha$ above the best candidate with probability at least
$1-\beta/2$.  A union bound over the training and validation failure
events proves \Cref{eq:err-Lstar}.

Finally, since $T\sqrt{D_p}=2D_p$ and $k=\bigO{\log(1/\beta)}$, the choices of $n_0$ and $m$ give the sample bound of \Cref{thm:improper-agnostic-gaussian}.
Computing the features, running the $k$ private optimizers, and
evaluating the $kM$ candidates on $S_{\mathrm{val}}$ take
$\operatorname{poly}(n)$ arithmetic operations.
The detailed proof, along with privacy and runtime calculations appear
in~\Cref{app:improper-gaussian}.
\looseness=-1

\begin{remark}
  \label{remark:all-dimensions}
  The dimension assumption in \Cref{thm:improper-agnostic-gaussian}
  allows us to use the clipping radius $T=2\sqrt{D_p}$.
  To obtain a guarantee in every dimension, we instead take
  $T=\sqrt{D_p/(c\alpha)}$
  for a sufficiently small universal constant $c>0$.
  Orthonormality and Markov's inequality give
  \begin{equation}
    \label{eq:universal-clipping}
    \Pr_\mu(\|\psi_p(X)\|_2>T)
    \le
    \frac{D_p}{T^2}
    =c\alpha.
  \end{equation}
  Replacing \Cref{lem:hermite-clipping} with this bound, the same
  analysis~(see \Cref{app:improper-gaussian}) establishes the learning
  guarantee of \Cref{thm:improper-agnostic-gaussian} for every
  $d\ge1$.  The statistical term in the sample complexity remains
  unchanged, and the privacy term acquires an additional
  $\alpha^{-1/2}$ factor.
\end{remark}

\section{Private Proper Agnostic Learning}
\label{sec:proper-gaussian}
The agnostic learner in \Cref{sec:improper} is improper since its
output need not be a halfspace. In this section, we provide a private
proper learner that outputs a halfspace in \(\reals^d\). A halfspace
requires only $\bigO{d}$ parameters, compared to
$d^{\bigO{1/\alpha^2}}$ coefficients for the improper hypothesis, and
is faster to evaluate on a point and also easier to interpret.

We will adapt the low-rank subspace reduction
of~\citet{diakonikolas2021proper} to design the learner. First, it
constructs a finite class of halfspaces from a polynomial with
sufficiently small excess squared loss. The resulting class is
guaranteed to contain a halfspace whose error is close to the best
halfspace. Once we have this finite class of halfspaces, we use the
exponential mechanism~\citep{mcsherry2007mechanism} to privately
select a halfspace from it. Our main result is stated below.

\begin{theorem}[DP proper agnostic learner]\label{thm:private-proper-gaussian}
Let $\alpha, \beta, \deltaDP \in (0, 1/2)$ and $\epsDP \in (0, 1)$.
Set $p = \lceil C_0/(c\alpha)^2 \rceil$ for a sufficiently large universal constant $C_0 > 0$
and a sufficiently small universal constant $c > 0$, let $D_p = \binom{d+p}{p}$, and suppose
that $d \ge C_{\mathrm{dim}}\,\alpha^{-5}$ for a sufficiently large universal constant
$C_{\mathrm{dim}} > 0$.
Let $\jointdist$ be any distribution on $\mathbb{R}^d \times \{\pm 1\}$ with marginal
$\xmarginal = \mathcal{N}(0, I_d)$. Then \Cref{alg:private-proper-gaussian} is
$(\epsDP, \deltaDP)$-DP and, given
\[
  n = \softO{\frac{D_p}{\alpha^2} + \frac{D_p}{\epsDP\,\alpha}
      + \frac{1}{\alpha^6} + \frac{1}{\epsDP\,\alpha^5}}
\]
i.i.d.\ samples from $\jointdist$, returns a halfspace $h \in \mathcal{C}$ satisfying
$\err[\jointdist]{h} \le \OPT + \alpha$ with probability at least $1 - \beta$.
The algorithm runs in time $\operatorname{poly}(n) \cdot \exp\bigl(O(\alpha^{-4}\log(1/\alpha))\bigr)$.
\end{theorem}

\noindent A proof sketch appears at the end of this section, and the full
proof is given in \Cref{app:proper-gaussian}.
We use a strengthened version of the subspace reduction
of~\citet{diakonikolas2021proper}, reducing the polynomial degree
from $\bigO{1/\alpha^4}$ to $\bigO{1/\alpha^2}$.
This reduces the subspace dimension from $\bigO{1/\alpha^6}$
to $\bigO{1/\alpha^4}$ and yields a smaller finite class for
the exponential mechanism.
We also extend the reduction to labels in $\bs{-1,1}$.
\Cref{lem:proper-finite-class} states this guarantee.

Thus, our sample bound has the same $\binom{d+p}{p}$ dependence
on the dimension as non-private improper agnostic learning
under Gaussian marginals~\citep{diakonikolas2010bounded},
while returning a halfspace.
When $\epsDP\ge\alpha$, the privacy terms are absorbed into
the statistical terms $D_p/\alpha^2+1/\alpha^6$, up to
logarithmic factors. In this regime, privacy comes essentially
for free.

The factor $\exp\bigl(\bigO{\alpha^{-4}\log(1/\alpha)}\bigr)$ in the running time arises from
constructing and searching a finite cover of halfspaces of at most this size, so the running
time is polynomial in $n$ for every fixed accuracy~$\alpha$. The non-private proper learner
of~\citet{diakonikolas2021proper} also searches such a cover, and its running time
likewise includes a term exponential in $\mathrm{poly}(1/\alpha)$.

\begin{lemma}
  \label{lem:proper-finite-class}
  There exist universal constants $C_0,C,c_0>0$ such that the following
  holds. Let $(X,U)$ have any joint distribution with
  $X\sim\cN(0,I_d)$ and $\abs{U}\le1$. Fix $\alpha\in(0,1/2)$,
  and set $p=\ceil{C_0\alpha^{-2}}$. Let $\cP_p$ denote the space of
  polynomials on $\reals^d$ of degree at most $p$. Suppose that
  $q\in\cP_p$ satisfies
  \begin{equation}\label{eq:sq-loss-bound}
    \bE\bs{\br{U-q(X)}^2}
    \le \inf_{Q\in\cP_p}\bE\bs{\br{U-Q(X)}^2}
    +c_0^2\alpha^2.
  \end{equation}
  Using just $q$ and $\alpha$, we can construct a finite class of
  halfspaces $\cH(q)\subseteq\cC$ satisfying \(\log\abs{\cH(q)}\le
  C\alpha^{-4}\log(1/\alpha),\) such that, for every $f\in\cC$, there
  exists $h\in\cH(q)$ with
  \(
  \bE\bs{U\br{f(X)-h(X)}}\le C\alpha.
  \)
\end{lemma}

\noindent The proof is stated in~\Cref{app:proper-gaussian}. A natural
idea is to obtain the polynomial \(q\) from our private improper
learner in \Cref{thm:improper-agnostic-gaussian}. However, the
coefficients $\widehat v_j$ obtained in each regression run of
\Cref{alg:private-polynomial-regression} only satisfy the following
$\Lone$ guarantee:
\[
  \bE_{(X,Y)\sim \jointdist}
  \abs{\ip{\widehat v_j}{\bar\psi_{p,T}(X)}-Y}
  \le \min_{\norm{v}_2\le1}\bE_{(X,Y)\sim \jointdist}
  \abs{\ip{v}{\bar\psi_{p,T}(X)}-Y}+\bigO{\alpha},
\]
which does not imply the excess squared-loss bound in
\Cref{eq:sq-loss-bound}.  Moreover, the function \(\ip{\widehat
  v_j}{\bar\psi_{p,T}(X)}\) is a linear combination of clipped Hermite
features \(\bar\psi_{p,T}(X)\) and therefore need not be a polynomial
as required in~\Cref{lem:proper-finite-class}.
We therefore need a different private polynomial construction that
directly controls excess squared loss.

\subsection{Constructing a private polynomial with small excess squared loss}

To construct the polynomial, we propose a simple estimator based on
the correlations between the labels and the Hermite features. For
squared-loss regression, these correlations give the coefficients of
the optimal polynomial~(see~\Cref{lem:hermite-squared-loss-minimizer}
below). However, clipping changes the correlations between the labels
and the Hermite features. Write $m=\bE\bs{Y\psi_p(X)}$ for the
original correlation vector and $m_T=\bE\bs{Y\bar\psi_{p,T}(X)}$ for
the correlation vector computed using clipped features. Then
\begin{align*}
  \norm{m-m_T}_2
   & =\sup_{\norm{v}_2=1}
  \abs{\bE\bs{Y\br{1-s(X)}\ip{v}{\psi_p(X)}}} \\
   & \le\sqrt{\bE\bs{\br{1-s(X)}^2}}
  \le\sqrt{\Pr_\xmarginal\bs{\norm{\psi_p(X)}_2>T}}
  =\bigO{\sqrt{\alpha}},
\end{align*}
where the first inequality uses Cauchy--Schwarz and orthonormality of
Hermite functions, and the last inequality uses $\bigO{\alpha}$ bound
on the clipping probability
from~\Cref{sec:improper}~(see~\Cref{lem:hermite-clipping}). Thus, a
\(\bigO{\sqrt{\alpha}}\) bound on the difference between the original
and clipped correlations only gives an \(\bigO{\alpha}\) bound on excess
squared loss, whereas~\Cref{lem:proper-finite-class} requires an
\(\bigO{\alpha^2}\) excess squared loss. Therefore, we need our estimated
coefficients to be within \(\bigO{\alpha}\) of the optimal coefficients.

\begin{algorithm}[t]
  \caption{Private proper learning of Gaussian halfspaces}
  \label{alg:private-proper-gaussian}
  \begin{algorithmic}[1]
    \Require Sample $S$ of size $n$, accuracy $\alpha$, failure probability $\beta$,
    and privacy parameters
    $(\epsDP,\deltaDP)$.
    \State Set $p=\ceil{C_0/(c\alpha)^2}$, $D_p=\binom{d+p}{p}$,
    and $T=2\sqrt{D_p}$, where $c>0$ is a sufficiently small
    universal constant.
    \label{step:proper-gaussian-degree}
    \State Split $S$ into disjoint blocks $S_1,S_2$ of size $n/2$ each,
    independently of the sample values.
    \label{step:proper-gaussian-parameters}
    \State Draw $Z\sim\cN(0,\sigma^2I_{D_p})$ with $\displaystyle
      \sigma=
      \frac{4T\sqrt{2\log(1.25/\deltaDP)}}
      {\epsDP n}$.
    \State $\displaystyle
      \widetilde m\gets
      \operatorname{Proj}_{B_2^{D_p}}\Biggl(
        \frac2n\sum_{(X_i,Y_i)\in S_1}
        Y_i\bar\psi_{p,T}(X_i)+Z\Biggr)$ and set
    $q(x)\gets\ip{\widetilde m}{\psi_p(x)}$.
    \label{step:proper-gaussian-regression}
    \label{step:proper-gaussian-polynomial}
    \State Construct $\cH(q)$ using the construction in
    \Cref{lem:proper-finite-class} with accuracy $c\alpha$.
    \label{step:proper-gaussian-cover}
    \State Sample $h\in\cH(q)$ with probability proportional to
    $\displaystyle
      \exp\Biggl(-(\epsDP/2)
        \sum_{(X_i,Y_i)\in S_2}\mathbf1\bc{h(X_i)\ne Y_i}\Biggr)$.
    \label{step:proper-gaussian-selection}
    \State \Return $h$.
  \end{algorithmic}
\end{algorithm}

We show that this can be achieved by constructing the polynomial for
the clipped labels $U=s(X)Y$, using the loss
$L_s(Q)=\bE\bs{\br{U-Q(X)}^2}$. Note that $\abs{U}=\abs{s(X)Y}\le 1$.
Then, the corresponding correlation vector is
\(m_T=\bE\bs{Y\bar\psi_{p,T}(X)} =\bE\bs{U\psi_p(X)}.\) By
orthonormality of the Hermite basis, $m_T$ is exactly the coefficient
vector minimizing $L_s$ over $\cP_p$.

\begin{lemma}
  \label{lem:hermite-squared-loss-minimizer}
  Let $X\sim\cN(0,I_d)$, let $\abs{U}\le1$, and write
  $m_T=\bE\bs{U\psi_p(X)}$. The polynomial
  $q_{m_T}(x)=\ip{m_T}{\psi_p(x)}$ is the unique minimizer of
  $L_s$ over $\cP_p$.
\end{lemma}

\begin{proof}[Proof of~\Cref{lem:hermite-squared-loss-minimizer}]
  Every polynomial in $\cP_p$ can be written as
  $q_w(x)=\ip{w}{\psi_p(x)}$ for some $w\in\reals^{D_p}$.  Using
  orthonormality of the Hermite basis, we get
  $\bE\bs{\psi_p(X)\psi_p(X)^\top}=I_{D_p}$. Then,
  \begin{equation}\label{eq:expand-sq-loss}\begin{aligned}
      L_s(q_w) = \bE\bs{\br{U-q_w(X)}^2} & =\bE\bs{U^2}-2\ip{w}{m_T}+\norm{w}_2^2        \\
                                         & =\bE\bs{U^2}-\norm{m_T}_2^2+\norm{w-m_T}_2^2.
    \end{aligned}
  \end{equation}
  The first two terms do not depend on $w$.  Thus, the loss is minimized
  uniquely at $w=m_T$.
\end{proof}
\noindent Thus, our goal is to estimate \(m_T\) privately. For any
estimate $\widetilde m$, using~\Cref{eq:expand-sq-loss}, we get that
the polynomial $q_{\widetilde m}(x)=\ip{\widetilde m}{\psi_p(x)}$
satisfies
\[
  L_s(q_{\widetilde m})-\inf_{Q\in\cP_p}L_s(Q)
  =\norm{\widetilde m-m_T}_2^2.
\]
It therefore suffices to estimate a coefficient vector
\(\widetilde{m}\) such that \(\norm{\widetilde{m}-m_T}_2\le
c_0c\alpha\) for some constant $c_0>0$. The regression step in
\Cref{alg:private-proper-gaussian}
(Step~\ref{step:proper-gaussian-regression}) estimates these correlations
using the Gaussian mechanism and projects the estimate onto the
coefficient unit ball. Since $\norm{m_T}_2\le1$, projection cannot
increase the estimation error. The following lemma gives the guarantee
for the resulting polynomial.

\begin{lemma}
  \label{lem:private-l2-polynomial}
  Let $\alpha,\beta,\deltaDP\in(0,1/2)$ and
  $\epsDP\in(0,1]$. Let $q$ be the polynomial
  constructed in Step~\ref{step:proper-gaussian-polynomial} of
  \Cref{alg:private-proper-gaussian}, using the $n/2$ independent
  samples in $S_1$. Its release is
  $(\epsDP,\deltaDP)$-differentially
  private, and $\Ltwonorm[\xmarginal]{q}\le1$ on every run. If
  $n/2=\softOm{D_p/\alpha^2+D_p/(\epsDP\alpha)}$,
  then, with probability at least $1-\beta/2$,
  \begin{equation}\label{eq:proper-bounded-label-objective}
    L_s(q)\le\inf_{Q\in\cP_p}L_s(Q)+c_0^2c^2\alpha^2.
  \end{equation} The proof is given in
  \Cref{app:proper-gaussian}.
\end{lemma}

\subsection{Constructing the finite set of halfspaces}
\label{subsec:proper-finite-halfspaces}

We now use the polynomial $q$ computed
in Step~\ref{step:proper-gaussian-polynomial} of
\Cref{alg:private-proper-gaussian} to construct a finite class of
halfspaces. On the event in \Cref{lem:private-l2-polynomial}, this
polynomial satisfies the squared-loss
condition~(\Cref{eq:sq-loss-bound}) of \Cref{lem:proper-finite-class}
with accuracy $c\alpha$. We therefore apply
\Cref{lem:proper-finite-class} in
Step~\ref{step:proper-gaussian-cover} to construct $\cH(q)$.  The full
algorithm and proof are given in \Cref{app:proper-gaussian}. The
algorithm follows the subspace and discretization construction
of~\citet{diakonikolas2021proper}. Specifically, we compute
\[
  M=\bE_{X\sim\cN(0,I_d)}\bs{\nabla q(X)\nabla q(X)^\top}
\]
and take $V$ to be the span of its eigenvectors with eigenvalues at
least $c^2\alpha^2/C_0$. They show that the resulting subspace has
dimension $\bigO{\alpha^{-4}}$. Discretizing unit normal vectors in
$V$ gives a finite class of halfspaces satisfying
$\log\abs{\cH(q)}=\bigO{\alpha^{-4}\log(1/\alpha)}$. All these
operations use only \(q\) and the known Gaussian marginal, so~\Cref{step:proper-gaussian-cover} preserves DP by
post-processing~(\Cref{lem:post-processing}).

\Cref{lem:private-l2-polynomial} guarantees the polynomial \(q\) has
small excess squared loss for the clipped labels $U$.
\Cref{lem:proper-class-accuracy} next translates this guarantee into
classification error for the original labels.

\begin{lemma}
  \label{lem:proper-class-accuracy}
  Suppose that $d\ge C_{\mathrm{dim}}\alpha^{-5}$ for a sufficiently
  large universal constant $C_{\mathrm{dim}}$ and $C$ is the constant in
  \Cref{lem:proper-finite-class}. Let $q$ and $\cH(q)$ be constructed by
  \Cref{alg:private-proper-gaussian}. If $q$ satisfies
  \Cref{eq:proper-bounded-label-objective},
  \[
    \min_{h\in\cH(q)}\err[\jointdist]{h}
    \le\OPT+(C/2+1)c\alpha.
  \]
\end{lemma}

\begin{proof} On the event in the lemma,
  \Cref{lem:proper-finite-class} gives, for every halfspace $f\in\cC$,
  an $h\in\cH(q)$ satisfying $\bE\bs{U\br{f(X)-h(X)}}\le Cc\alpha$.
  Since
  \begin{align*}
    \err[\jointdist]{h}-\err[\jointdist]{f}
     & =\tfrac12\bE\bs{Y\br{f(X)-h(X)}}\le\tfrac12\bE\bs{U\br{f(X)-h(X)}}+\bE\bs{1-s(X)},
  \end{align*}
  it remains to bound the last term. As $s(X)=1$ outside the clipping
  event, \Cref{lem:hermite-clipping} gives
  \(
  \bE\bs{1-s(X)}
  \le\bP\bs{\norm{\psi_p(X)}_2>T}\le c\alpha
  \) when $T=2\sqrt{D_p}$ and $d\ge2p^2/(c\alpha)$.\end{proof}

\noindent Thus, $\cH(q)$ contains a halfspace with error at most
$\OPT+(C/2+1)c\alpha$.

\subsection{Selecting the best halfspace}

Finally,~\Cref{step:proper-gaussian-selection} selects a halfspace
from $\cH(q)$ using the exponential mechanism, with scores equal to
the negative empirical errors on $S_2$. These errors are measured
against the original labels $Y$.  Conditional on the released
polynomial, $\cH(q)$ is fixed and independent of $S_2$.  By
\Cref{lem:private-finite-validation}, this selection adds at most
$c\alpha$ error with probability at least $1-\beta/2$, provided
\begin{equation}\label{eq:proper-validation-size}
  n=\Om{
    \br{\alpha^{-4}\log(1/\alpha)+\log(1/\beta)}
    \br{\frac1{\alpha^2}
      +\frac1{\epsDP\alpha}}}.
\end{equation}
Combining this with \Cref{lem:private-l2-polynomial} and taking $c$
sufficiently small gives error at most $\OPT+\alpha$, with
probability at least $1-\beta$. The private regression step uses
$S_1$, while the exponential mechanism uses the disjoint block $S_2$,
each of size $n/2$. Therefore, post-processing and parallel
composition give
$(\epsDP,\deltaDP)$-privacy for
\Cref{alg:private-proper-gaussian}.

\begin{proof}[Proof Sketch of~\Cref{thm:private-proper-gaussian}]
  \Cref{alg:private-proper-gaussian} is the proper private learner
  achieving the guarantee in \Cref{thm:private-proper-gaussian}.
  Its output belongs to $\cH(q)\subseteq\cC$, so it is proper.
  \Cref{lem:private-l2-polynomial} guarantees
  that~\Cref{step:proper-gaussian-polynomial} outputs a polynomial
  satisfying \Cref{eq:sq-loss-bound} in \Cref{lem:proper-finite-class}
  with accuracy $c\alpha$, with probability at least $1-\beta/2$.  Thus,
  Step~\ref{step:proper-gaussian-cover} constructs $\cH(q)$ from a
  subspace of dimension
  $\bigO{\alpha^{-4}}$.~\Cref{lem:proper-class-accuracy} ensures that it
  contains a halfspace with error at most
  $\OPT+(C/2+1)c\alpha$.  When the sample size $n$ satisfies
  \Cref{eq:proper-validation-size}, the exponential mechanism adds at
  most $c\alpha$ error with probability at least $1-\beta/2$ by
  \Cref{lem:private-finite-validation}; a union bound and sufficiently
  small $c$ give the claimed accuracy. Adding the validation sample
  cost to the sample cost in
  \Cref{lem:private-l2-polynomial} gives the stated sample complexity.

  For privacy, note that
  Step~\ref{step:proper-gaussian-regression} is
  $(\epsDP,\deltaDP)$-DP by
  \Cref{lem:private-l2-polynomial}, Step~\ref{step:proper-gaussian-cover} is
  post-processing, and Step~\ref{step:proper-gaussian-selection} is
  $\epsDP$-DP on $S_2$ by exponential mechanism,
  conditioned on the released polynomial. Since $S_1$ and $S_2$ are
  disjoint, parallel composition makes the entire algorithm
  $(\epsDP,\deltaDP)$-DP.
\end{proof}

\section{Private Proper Semi-Agnostic Learning}
\label{sec:semi-agnostic-proper-gaussian}
In this section, we provide a DP algorithm for proper semi-agnostic
learning of halfspaces whose sample complexity is polynomial in $d$
and $1/\alpha$, rather than exponential in $1/\alpha$ as
in~\Cref{thm:private-proper-gaussian,thm:improper-agnostic-gaussian}.
The difference between semi-agnostic and agnostic error guarantee is
that the former allows a universal constant factor multiplying $\OPT$.
We use a privatised version of the non-private gradient descent~(on a
non-convex function) algorithm of~\citet{DiakonikolasKTN22} as a
subroutine. Roughly speaking, given an upper bound $\eta\geq\OPT$,
this private subroutine~(\Cref{alg:semi-agnostic-subroutine}) returns
a halfspace with error $\bigO{\eta}$. Then, our main
algorithm~(\Cref{alg:semi-agnostic}) searches over a small set of
candidate values of $\eta$ and selects a hypothesis using the
exponential mechanism.  We state our main result below and defer the
privacy proof and technical details of the accuracy analysis
to~\Cref{app:semi-agnostic}.

\begin{theorem}[DP semi-agnostic proper learner]\label{thm:semi-agnostic}
Let $\alpha, \beta, \epsDP, \deltaDP \in (0, 1)$.
Let $\jointdist$ be any distribution on $\mathbb{R}^d \times \{\pm 1\}$ with marginal
$\xmarginal = \mathcal{N}(0, I_d)$. Then \Cref{alg:semi-agnostic} is
$(\epsDP, \deltaDP)$-DP and, given\looseness=-1
\[
  n = \softO{\frac{d}{\alpha^2} + \frac{d}{\varepsilon_{\mathrm{DP}}\,\alpha}}
\]
i.i.d.\ samples from $P$, where $d \ge 16\log(n/\beta)$, it returns a halfspace
$h \in \mathcal{C}$ satisfying \(
   \err[\jointdist]{h}
   \leq C\cdot\OPT+\alpha\)
with probability at least $1 - \beta$, where $C > 0$ is a universal constant.
The algorithm runs in time $\operatorname{poly}(n)$.
\end{theorem}

We make the above assumption on $d$ (\Cref{sec:semi-agnostic-reg-cond}) for ease of exposition. It can be removed by setting the clipping parameter $L$ in \Cref{alg:semi-agnostic-subroutine} to a tighter bound of $O(\sqrt{d}+\sqrt{\log(n/\beta)})$, thus simply introducing more $\operatorname{polylog}$ terms in the sample complexity that we already omit in our presentation.\looseness=-1

\subsection{The Algorithm}
We first describe the subroutine
in~\Cref{alg:semi-agnostic-subroutine}, which takes a known upper
bound $\eta\geq\OPT$ and performs gradient updates using a ramp loss.
For $a>0$ and $b\in\reals$, define
\begin{equation}\label{eq:semi-agnostic-ramp}
   r_{a,b}(z)=
   \begin{cases}
      0     & \text{if } z\leq b-a, \\
      z-b+a & \text{if } b-a<z<b+a, \\
      2a    & \text{if } z\geq b+a.
   \end{cases}
\end{equation}

\begin{algorithm}[ht!]
   \caption{DP Semi-Agnostic Subroutine $\DPSAS$}\label{alg:semi-agnostic-subroutine}
   \begin{algorithmic}[1]
      \Require Dataset $D$ of $n_0$ labelled examples,
      privacy parameters $\epsDP,\deltaDP>0$, and error parameter $\eta$.

      \LeftComment{Partition Dataset and Clip for DP}
      \State $s \gets \Theta(1/\log^{3/2}(1/\eta))$ and $T \gets \Theta(\log^4(1/\eta))$.
      \State Partition $D$ into $D_{init},D_\tau,D_0,\dots,D_{T-1}$, each of size $m_0 \gets |D|/(T+2)$.
      \State $L \gets 2\sqrt{d}$ and clip each feature vector in $D$ to have norm at most $L$.\label{step:semi-agnostic-clipping}

      \LeftComment{Privately computing threshold $\hat{\tau}$.}
      \State $\kappa \gets C\eta$ for a universal constant $C$ and $\hat{y} \gets \mathbb{E}_{(x,y) \sim D_\tau}[y] + z_\tau$,\label{step:semi-agnostic-threshold}
      \Statex \hspace{\algorithmicindent} where $z_\tau \sim \Lap\left(\tfrac{2}{\epsDP m_0}\right)$.
      \State $\hat{\tau} \gets \Phi^{-1}\left(\mathsf{clip}\left[\tfrac{1+\hat{y}}{2}\right]_\kappa^{1-\kappa}\right)$.

      \LeftComment{Privately computing initial weight $\hat{w}_0$.}
      \State $\tilde{w}_0 \gets \mathbb{E}_{(x,y) \sim D_{init}}[xy] + z_0$,\label{step:semi-agnostic-initialization}
      \Statex \hspace{\algorithmicindent} where $z_0 \sim \mathcal{N}(0,\sigma^2\mathbb{I})$ and $\sigma \gets \tfrac{2L\sqrt{2\log(2/\deltaDP)}}{\epsDP m_0}$.
      \State $\hat{w}_0 \gets \tfrac{\tilde{w}_0}{\|\tilde{w}_0\|}$ if $\|\tilde{w}_0\| \neq 0$, otherwise $\hat{w}_0 \sim \mathcal{S}^{d-1}$ uniformly at random.

      \LeftComment{Run DP online gradient descent.}
      \State $\lambda \gets e^{\hat{\tau}^2/2}s$ and $B \gets \mathcal{S}^{d-1}$.
      \For{$t = 0$ to $T-1$}
      \State $\phi_t \gets (\pi/2)(1-s^2/64)^t$, $a_t \gets \sin(\phi_t)$, and $b_t \gets -\hat{\tau}\cos(\phi_t)$.
      \State $\hat{\mathcal{L}}_t(w) \gets \mathbb{E}_{(x,y)\sim D_{t}}\left[-r_{a_t,b_t}\left(\tfrac{w\cdot x}{\|w\|}\right)y\right]$.
      \State $\hat{u} \gets \hat{w}_t - \lambda(\nabla \hat{\mathcal{L}}_t(\hat{w}_t)+z_{t+1})$ and $\hat{w}_{t+1} \gets \textrm{proj}_B(\hat{u})$,\label{step:semi-agnostic-gradient}
      \Statex \hspace{\algorithmicindent} where $z_{t+1} \sim \mathcal{N}(0,\sigma^2(\mathbb{I}-\hat{w}_t \hat{w}_t^\top))$.
      \EndFor

      \State \textbf{Return} $\hat{w}_T,\hat{\tau}$.

   \end{algorithmic}
\end{algorithm}

\Cref{alg:semi-agnostic-subroutine} follows Algorithms~1 and~2
from~\citet{DiakonikolasKTN22}, with noise added to the computations
that access the dataset. The algorithm needs to privately estimate two
quantities from the dataset before running gradient descent: the
threshold parameter $\tau$ and the initial weight vector
$\hat{w}_0$.~\citet{DiakonikolasKTN22} first estimate the threshold
from the average of the labels in $D_\tau$, using the inverse Gaussian
CDF. We privatise this average by adding Laplace noise, obtaining the
threshold estimate $\hat{\tau}$
in~\Cref{step:semi-agnostic-threshold}.  Next, we compute the initial
weight vector $\hat{w}_0$ by adding Gaussian noise to the average of
$xy$ over $D_{init}$ in~\Cref{step:semi-agnostic-initialization}. The
algorithm then keeps $\hat{\tau}$ fixed, and performs gradient descent
using the ramp function in~\Cref{eq:semi-agnostic-ramp} and the private threshold $\hat{\tau}$, with a fresh
block $D_t$ in each iteration. \Cref{step:semi-agnostic-gradient} adds Gaussian noise to
each empirical gradient before performing the gradient update.

With these modifications, \Cref{alg:semi-agnostic-subroutine}
privately returns a hypothesis with population classification error
$\bigO{\eta}$ whenever $\eta\geq\OPT$, as shown
in~\Cref{prop:semi-agnostic-subroutine}. Since $\OPT$ is unknown,
\Cref{alg:semi-agnostic} tries a geometric grid of candidate values
$\eta_1,\dots,\eta_K$. For each $\eta_k$, it runs
\Cref{alg:semi-agnostic-subroutine} on a separate subset of the data
to obtain a hypothesis $h_k$. The grid is chosen such that at least
one of these hypotheses has error $\bigO{\OPT+\alpha}$ with high
probability. The algorithm then uses the exponential mechanism on a
separate validation set $D_{val}$ to select a hypothesis whose
empirical classification error is close to the smallest among these
candidates.

\subsection{Accuracy}\label{sec:semi-agnostic-acc}

A major component for proving the utility for~\Cref{alg:semi-agnostic}
is proving the accuracy guarantees
for~\Cref{alg:semi-agnostic-subroutine}. Note that by construction,
$\alpha \leq \eta_k$ for all $k \in [K]$. In the presentation for the
rest of~\Cref{sec:semi-agnostic-acc}, we will omit the subscript from
$\eta_k$ to refer to the parameter $\eta$
in~\Cref{alg:semi-agnostic-subroutine}.
We write $m_0$ for the number of examples in each of the
$T+2$ disjoint blocks used by the algorithm; the total sample
size for one run is $n_0=(T+2)m_0$.

\begin{proposition}\label{prop:semi-agnostic-subroutine}
Let $\eta \in (0, 1/2)$ and $\beta, \varepsilon_{\mathrm{DP}}, \delta_{\mathrm{DP}} \in (0, 1)$.
Let $\jointdist$ be any distribution on $\mathbb{R}^d \times \{\pm 1\}$ with marginal
$\xmarginal = \mathcal{N}(0, I_d)$ and $\OPT \le \eta$. Then
\Cref{alg:semi-agnostic-subroutine} is $(\epsDP, \deltaDP)$-DP and, given
\[
      m_0=\softO{
         \frac{d}{\eta^2}
         +\frac{d}{\epsDP\eta}}
   \]
i.i.d.\ samples from $\jointdist$ in each of its blocks, where $d \ge 16\log(n_0/\beta)$ for the
total sample size $n_0$, returns a halfspace $h \in \mathcal{C}$ satisfying
$\err[\jointdist]{h}\leq C\eta$ with probability at least $1 - \beta$, where $C > 0$ is a
universal constant.
\end{proposition}

We follow the utility analysis of~\citet{DiakonikolasKTN22}, with
modifications to account for the noise added for privacy. Their
analysis starts with a weight vector that makes an angle less than
$\pi/2$ with the optimal halfspace $w^\star$. In each iteration, the
gradient update reduces this angle until the weight vector is
sufficiently close to \(w^\star\) to give the desired error guarantee.
We first analyse these updates assuming that our private
initialization satisfies this starting condition and that replacing
the optimal threshold by our private estimate preserves a small
classification and then prove it
in~\Cref{clm:semi-agnostic-initialization,clm:semi-agnistic-threshold}.
They first prove \emph{population-level} results, where they show that
applying population-level gradients will slowly lead us to $w^\star$,
and then use concentration bounds to show that the empirical gradients
satisfy the estimates needed for convergence with high probability. We
modify these empirical estimates to account for the additional noise
added for privacy.

The first result describes the structure of the noisy gradient and the
updated weight vector. Its proof is given in~\Cref{app:semi-agnostic}.
\begin{claim}[Modification of Claim~C.2 from~\citet{DiakonikolasKTN22}]\label{clm:noisy-grad-def}
   For any unit vector $w \in \mathbb{R}^d$, the DP gradient rule from~\Cref{alg:semi-agnostic-subroutine} with loss $\hat{\mathcal{L}}_t(w)$ and stepsize $\lambda$ corresponds to the update:
   $$w' = \frac{w - \lambda\nabla_w \tilde{\mathcal{L}}_t(w)}{\|w - \lambda\nabla_w \tilde{\mathcal{L}}_t(w)\|},$$
   such that $\nabla_w \tilde{\mathcal{L}}_t(w) = \mathbb{E}_{(x,y) \sim D_t}\left[-y \mathbf{1}\{|w\cdot x -b_t| \leq a_t\} \nabla_w\tfrac{w\cdot x}{\|w\|}\right] +\, z_{t+1}$, where $z_{t+1} \sim \mathcal{N}\left(0,(\mathbb{I}-ww^\top)\sigma^2\right)$ and $\sigma = \tfrac{2L\sqrt{2\log(2/\deltaDP)}}{\epsDP m_0}$.
\end{claim}

Next, we prove the DP analogue of Lemma~C.12 from~\cite{DiakonikolasKTN22} and show that the original additive error guarantees are not affected by the added noise for DP. We keep the notation consistent with theirs, and define $g(w,\phi,x,y) = y (\textrm{proj}_{w^\bot}x)\mathbf{1}\{|w\cdot x + \hat{\tau}\textrm{cos}(\phi)| \leq \textrm{sin}(\phi)\}$ and $B_{\phi} = \{z \in \mathbb{R}: |z+\hat{\tau}\textrm{cos}(\phi)| \leq \textrm{sin}(\phi)\}$. Our GD algorithm outputs $f(w,\phi,x,y) \coloneqq \mathbb{E}_{(x,y) \sim D}[g(w,\phi,x,y)] + z$, where $z$ is the Gaussian noise (as in~\Cref{clm:noisy-grad-def}).
\begin{lemma}[Modification of Lemma~C.12 from~\citet{DiakonikolasKTN22}]\label{lem:semi-agnostic-gradient-update}
   Let $q=\Pr_{x \sim \xmarginal}[w\cdot x \in B_{\phi}]$. Assuming that
   $$m_0 \geq \bigO{\frac{dq\log(1/\beta_0)}{\eta^2} + \frac{\log(d/\beta_0)}{q^2} + \frac{d\sqrt{\log(1/\deltaDP)\log(1/\beta_0)}}{\eta \epsDP}},$$
   we have with probability at least $1-\beta_0$ that for any unit vector $u \in \mathbb{R}^d$, such that $\mathbb{E}_{(x,y) \sim \jointdist}[g(w,\phi,x,y)\cdot u] \geq 0$, the following hold:
   $$f(w,\phi,x,y)\cdot u \geq \frac{1}{2} \mathbb{E}_{(x,y) \sim \jointdist}[g(w,\phi,x,y)\cdot u] - \bigO{\eta}$$
   and
   $$\|f(w,\phi,x,y)\| \leq 2 \|\mathbb{E}_{(x,y) \sim \jointdist}[g(w,\phi,x,y)]\| + \bigO{\eta}.$$
\end{lemma}
\begin{proof}
   The proof remains the same, except for accounting for the error due to Gaussian noise. This lemma is applied on independent dataset $D_t$ in each iteration $t$. We use the notation $D$ in this case as a shorthand for $D_t$, since we are omitting the subscript everywhere. For the first part, \citet{DiakonikolasKTN22} show the first line of the following.
   \begin{align*}
      u\cdot \mathbb{E}_{(x,y) \sim D}[g(w,\phi,x,y)] & \geq u\cdot \mathbb{E}_{(x,y) \sim \jointdist}[g(w,\phi,x,y)]\frac{\Pr_{x \sim D_x}[w\cdot x \in B_{\phi}]}{\Pr_{x \sim \xmarginal}[w\cdot x \in B_{\phi}]} - \bigO{\eta} \Pr_{x \sim \xmarginal}[w\cdot x \in B_{\phi}]         \\
      \implies u\cdot f(w,\phi,x,y)                   & \geq u\cdot \mathbb{E}_{(x,y) \sim \jointdist}[g(w,\phi,x,y)]\frac{\Pr_{x \sim D_x}[w\cdot x \in B_{\phi}]}{\Pr_{x \sim \xmarginal}[w\cdot x \in B_{\phi}]} - \bigO{\eta} \Pr_{x \sim \xmarginal}[w\cdot x \in B_{\phi}] - \|z\|
   \end{align*}
   Our private sample complexity allows $\|z\| \leq \bigO{\eta}$ with probability at least $1-\beta_0/2$ (via concentration of high-dimensional Gaussians). Therefore, this direction of the proof follows by following the remaining steps of the proof from~\citet{DiakonikolasKTN22}.

   We justify the norm bound separately. Let
   \(g_{\mathrm{pop}}\coloneqq\mathbb{E}[g(w,\phi,X,Y)]\) and \(\widehat g\coloneqq\mathbb{E}_{(x,y)\sim D}[g(w,\phi,x,y)]\),
   so that \(f=\widehat g+z\). Also, define the population and empirical masses of the band \(B_\phi\) by \(q\coloneqq\Pr_{X\sim \xmarginal}[w\cdot X\in B_\phi]\) and \(\widehat q\coloneqq\frac1n\sum_{i=1}^n
   \mathbf{1}\{w\cdot x_i\in B_\phi\}\), respectively.
   If \(\widehat g=0\), the claim is immediate. Otherwise, define
   \(
   u\coloneqq\widehat g/|\widehat g|.
   \)
   On the same concentration event on $z_\tau$ used above, and the argument in~\cite{DiakonikolasKTN22}, writing \(\widehat q/q\in[1/2,2]\), their argument gives, for every unit vector \(u\),
   \[ u\cdot\widehat g \le \frac{\widehat q}{q}\,u\cdot g_{\mathrm{pop}}+\bigO{\eta}.
   \]
   If \(\widehat g=0\), the desired bound is immediate. Otherwise, take \(u=\widehat g/\|\widehat g\|\).
   If \(u\cdot g_{\mathrm{pop}}\ge0\), then
   \[
      \|\widehat g\| \le 2\,u\cdot g_{\mathrm{pop}}+\bigO{\eta} \le 2\|g_{\mathrm{pop}}\|+\bigO{\eta}.
   \]
   If \(u\cdot g_{\mathrm{pop}}<0\), then, since \(\widehat q/q\ge1/2\),
   \[
      \frac{\widehat q}{q}\,u\cdot g_{\mathrm{pop}} \le \frac12\,u\cdot g_{\mathrm{pop}}<0,
   \]
   and hence \(\|\widehat g\|\le \bigO{\eta}\).
   Thus, in either case \(\|\widehat g\|\le2\|g_{\mathrm{pop}}\|+\bigO{\eta}\).
   The bound \(\|z\|\le \bigO{\eta}\) implies
   \[
      \|f\| \le \|\widehat g\|+\|z\| \le 2\|g_{\mathrm{pop}}\|+\bigO{\eta},
   \]
   as claimed.
\end{proof}

Now, writing \(\theta=\theta(w,w^\star)\), we get
\(v\cdot \mathbb{E}_{(x,y) \sim \jointdist}[g(w,\phi,x,y)] \ge \sin(\theta)e^{-t^2/2}>0\) using Lemma~C.3 from~\cite{DiakonikolasKTN22}.
They then condition on the event $A_k$ on which Lemma~C.12 holds, and apply the directional result of Lemma~C.12 to $v$. Hence, the only direction needed in their argument satisfies the positivity condition ($\mathbb{E}_{(x,y) \sim \jointdist}[g(w,\phi,x,y) \cdot u] \geq 0$) required by our restricted statement. This lower bound on
\(
v\cdot \mathbb{E}_{(x,y) \sim \jointdist}[g(w,\phi,x,y)]
\)
is then converted into a lower bound on
\(
w^\star\cdot \mathbb{E}_{(x,y) \sim \jointdist}[g(w,\phi,x,y)],
\)
and, together with the norm bound from Lemma~C.12, is used to verify the hypotheses of Corollary~C.8 and get contraction of the angle between the iterate and $w^*$.

Next, we show that the DP initialization $w_0$ in our algorithm satisfies the constraints that are needed in the proof of Theorem~1.2 from~\citep{DiakonikolasKTN22}. They require that the dot product of the true weight $w^*$ with our initialization be strictly more than $0$.
\begin{claim}\label{clm:semi-agnostic-initialization}
   Suppose
   \(
   \frac{e^{-\hat\tau^2/2}}{|\hat\tau|}\ge c\eta.
   \)
   Then, with probability at least \(1-\bigO{\beta}\), if
   \[
      m_0\ge \softO{\frac{d}{\eta^2}+\frac{\sqrt{d\log(1/\deltaDP)}}{\epsDP\eta}},
   \]
   the private initializer satisfies \(w^\star\cdot \widehat w_0>0\).
\end{claim}

We only invoke~\Cref{clm:semi-agnostic-initialization} in the above non-trivial regime. In the complementary case \(\frac{e^{-\hat\tau^2/2}}{|\hat\tau|}< C\eta\),
the argument of~\cite{DiakonikolasKTN22} already implies that using the threshold \(\hat\tau\) with any unit normal vector yields classification error \(\bigO{\eta}\). Hence, no positive-correlation initialization guarantee is needed in this case.
Finally, we bound the error on our noisy empirical threshold.
\begin{claim}[Modification of Claim~C.9 from~\citep{DiakonikolasKTN22}]\label{clm:semi-agnistic-threshold}
   Suppose $\eta,\beta_0 \in (0,1/2)$, and $w^*$ is a unit vector and $\tau$ is a threshold, such that $\Pr_{(x,y)\sim \jointdist}[\textrm{sign}(w^* \cdot x + \tau) \neq y] \leq \eta$. Suppose $\hat{\tau} = \Phi^{-1}\left(\mathsf{clip}[(1+\mathbb{E}_{(x,y)\sim D}[y] + z_\tau)/2]_\kappa^{1-\kappa}\right)$ for $\kappa=\bigO{\eta}$ is the DP estimate of the threshold $\tau$ output by~\Cref{alg:semi-agnostic-subroutine}. If
   $$m_0 \geq \bigO{\log(1/\beta_0)/\eta^2 + \log(1/\beta_0)/\epsDP\eta},$$
   then with probability at least $1-\beta_0$,
   $$\Pr_{(x,y) \sim \jointdist}[\textrm{sign}(w^* \cdot x + \hat{\tau}) \neq y] \leq \bigO{\eta}.$$
\end{claim}
\begin{proof}
   \cite{DiakonikolasKTN22} show that satisfying
   \(\left|\mathbb{E}_{(x,y)\sim D}[y]-\mathbb{E}_{(x,y)\sim \jointdist}[y]\right| \le \bigO{\eta}\) in the proof of Claim~C.9 is enough, which is what we need, as well. Thus,
   \(\left|\mathbb{E}_{(x,y)\sim D}[y] + z_\tau -\mathbb{E}_{(x,y)\sim \jointdist}[y]\right| \le \bigO{\eta}\)
   suffices for us, and the rest of the argument follows.
   We additionally clip the argument of $\Phi^{-1}$ to
   $[\kappa,1-\kappa]$, where $\kappa=C\eta$ for a sufficiently
   small universal constant $C>0$.

   Let \(\widetilde p = (1+\mathbb{E}_{(x,y)\sim \widehat D}[y]+z_\tau)/2\) and \(\widehat p = \mathsf{clip}\!\left(\widetilde p,[\kappa,1-\kappa]\right)\).
   Since $z_\tau$ is sampled from Laplace distribution with parameter $2/m_0\epsDP$, by the concentration of Laplace distribution, $|z_\tau| \leq 2\log(1/\beta_0)/m_0\epsDP$ with probability at least $1-\beta_0$. For this to be upper bounded by $\bigO{\eta}$, having $m_0 \geq \bigO{\log(1/\beta_0)/\epsDP\eta}$ suffices.
   For any $p\in[0,1]$,
   \(|\widehat p-p| \le |\widetilde p-p|+\kappa\).
   Hence, the clipping introduces at most an additional $\bigO{\eta}$ error. In particular, taking \(p=(1+\mathbb{E}_{(x,y)\sim \jointdist}[y])/2\), the above concentration guarantee implies \(|\widehat p-p|\le \bigO{\eta}\).
   Since $\widehat\tau=\Phi^{-1}(\widehat p)$, the remainder of the proof of Claim~C.9 follows as before.
\end{proof}

\subsection{Putting It All Together}

\begin{proof}[Proof of~\Cref{prop:semi-agnostic-subroutine}]
   As mentioned earlier, our modifications are needed only for dataset-level results. Claims~C.2 and~C.9, along with Lemma~C.12, are the only ones that needed modifications. Note that the technical results in Appendices~B and~C from~\cite{DiakonikolasKTN22} still hold when we are using independent samples for threshold estimation, initialization, and each iteration of gradient descent, so they transfer to our private case, as well. We have shown via~\Cref{clm:noisy-grad-def,clm:semi-agnistic-threshold}, and~\Cref{lem:semi-agnostic-gradient-update}, respectively, that at minimal cost in sample complexity, those results hold. Additionally, they prove that their initial weight vector $w_0$ is not too far in terms of angle from $w^*$. We prove in~\Cref{clm:semi-agnostic-initialization} that the same holds for our noisy initialization $\hat{w}_0$.
   With these necessary modifications, our proof of accuracy is complete.
\end{proof}

The grid in~\Cref{alg:semi-agnostic} starts at
$\eta_1=C_\eta\alpha$, where $C_\eta>1$ is a universal
constant, and uses $\eta_k=2^{k-1}\eta_1$ for $k\in[K]$.
The dataset is partitioned into a validation set $D_{val}$
and $K$ subsets $D^1,\dots,D^K$. For each $k\in[K]$,
the subroutine is run on $D^k$ with parameter $\eta_k$
to obtain $h_k$.

\begin{proof}[Proof of~\Cref{thm:semi-agnostic}]
   \Cref{prop:semi-agnostic-privacy} provides the privacy guarantees. Now, we complete the proof of accuracy. First, note that the difference between the final error guarantee required to be in $\bigO{\alpha}$, but in~\Cref{alg:semi-agnostic}, we work with $\eta_k$'s. The idea is that $\eta_k \geq \alpha$ for all $k \in [K]$. Now, either $\OPT < \eta_1$, or there must be a $k^*\in [K] \setminus \{1\}$, such that $\eta_12^{k^*-2} \leq \OPT \leq \eta_12^{k^*-1}$. We claim that in either case, there exists at least one $k \in [K]$, for which $h_k$ has error $\bigO{\OPT + \alpha}$.

   When $\OPT < \eta_1$, then~\Cref{alg:semi-agnostic-subroutine} will return $h$, such that the error of $h$ is at most $\bigO{\eta_1} = \bigO{\alpha}$ with probability at least $1-\beta'$, where $\beta'=\beta/K$ (\Cref{prop:semi-agnostic-subroutine}). This satisfies the semi-agnostic learning guarantee even if $\OPT$ is much smaller than $\eta$ because $\alpha = \Theta(\eta_1)$.

   On the other hand, suppose there exists $k^*\in [K] \setminus \{1\}$, such that $\eta_12^{k^*-2} \leq \OPT \leq \eta_12^{k^*-1}$. Then with probability at least $1-(K-1)\beta'$, for all $k \geq k^*$, \Cref{alg:semi-agnostic-subroutine} will return hypotheses $h_k$, respectively, such that the total error is $\bigO{\eta_{k}}$. Thus for $k^*$, $\eta_{k^*} = \eta_12^{k^*-1} \leq 2\OPT$. This means that $h_{k^*}$ should satisfy the semi-agnostic error guarantee $\bigO{\OPT}$ with no additional additive error.

   Combining the two together, with probability at least $1-\bigO{\beta}$, there must exist $k \in [K]$, such that the semi-agnostic error of $h_k$ is at most $\max\{\bigO{\alpha},\bigO{\OPT}\}$. Now, we just need to show that the final hypothesis returned by \Cref{alg:semi-agnostic} is accurate. For that, we know that with probability at least $1-\bigO{\beta}$, for all $k \in [K]$, the population error $\err[\jointdist]{h_k}$ must be close to the empirical error $\herr[D_{val}]{h_k}$ on $D_{val}$ using Hoeffding's inequality; because of our sample complexity,
   \[
      |\err[\jointdist]{h_k} - \herr[D_{val}]{h_k}| \leq \bigO{\sqrt{\frac{\log(K/\beta)}{|D_{val}|}}} \leq \bigO{\alpha}.
   \]
   With probability at least $1-\bigO{\beta}$, the exponential mechanism returns an $\bar{h}$, whose empirical error is at most $\bigO{\alpha}$ in addition to the empirical optimal among all $h_k$'s. From these two additional error sources, we get that $\bar{h}$ has a population error of at most $\max\{\bigO{\alpha},\bigO{\OPT}\} + \bigO{\alpha}$.

   Finally, we bound the running time.
   Each run of \Cref{alg:semi-agnostic-subroutine} uses every sample in its data a
constant number of times (for clipping, the estimates of $\hat{\tau}$ and $\hat{w}_0$, or
one gradient step), at $\bigO{d}$ cost per use. Since the runs use disjoint data, all
$K = \bigO{\log(1/\alpha)}$ runs together take $\bigO{nd}$ time. Scoring the $K$ candidates on
$D_{val}$ takes $\bigO{nd\log(1/\alpha)}$ time. Since $d \le n$, the total running
time is $\bigO{nd\log(1/\alpha)} = \operatorname{poly}(n)$.
\end{proof}

\section{Conclusion and Open Problems}
\label{sec:conclusion}
We provide the first computationally efficient DP agnostic learners for halfspaces, where privacy comes \emph{for free} in terms of sample complexity, i.e., the sample complexity overhead due to privacy matches that of the non-private learners. Specifically, we study three settings: improper agnostic, proper agnostic, and semi-agnostic. Our techniques are based on those of the non-private work, albeit with additional steps and analyses to incorporate clipping and noise injection to provide DP guarantees. \looseness=-1

One avenue for improvement is in the improper agnostic setting, where we have to explicitly compute the orthonormal basis for each family of distributions when the marginals are known, unlike~\citet{kalai2008agnostically}, who work with the standard polynomial basis for all families of distributions. The only parameter they need to know for their kernel is the degree of the polynomial. An interesting future direction is to find more versatile polynomial bases that admit
bounded coefficients for a large family of distributions, or alternatively to learn such
a basis privately from the data, removing the need to know the marginal in advance.
Another direction is to design efficient DP algorithms for agnostic learning of concept classes beyond halfspaces. Finally, one could explore lower bounds in DP agnostic learning to determine the necessary sample complexity overhead for these problems.

\section*{Acknowledgements and AI Disclosure}
The authors acknowledge the Novo Nordisk Foundation for support via the Startup grant (NNF24OC0\allowbreak087820); AS additionally acknowledges support from VILLUM FONDEN via the Young Investigator program (VIL72069).

The main ideas underlying this work were developed by the authors.
AI tools were used to help sharpen the parameter dependence of technical lemmas.
The authors take full responsibility for the correctness of the results and the content of the final manuscript.

\bibliographystyle{plainnat}
\bibliography{references}

\appendix
\crefalias{section}{appendix}
\crefalias{subsection}{subappendix}

\section{Other Related Work}
\label{app:related-wrok}
\paragraph{Agnostic learning under structured marginals.}
Without distributional assumptions, agnostically learning halfspaces is computationally hard, even for weak learning~\citep{guruswami2009hardness,feldman2006new,daniely2016complexity,tiegel2023hardness}.
Building on the low-degree Fourier approach of \citet{linial1993constant}, \citet{kalai2008agnostically} use polynomial $L_1$ regression to obtain improper agnostic learners for halfspaces with error $\OPT+\alpha$ under structured marginals, including Gaussian and log-concave distributions.
The complexity of this approach is governed by the degree of polynomials needed to approximate halfspaces in $L_1$. 
Under Gaussian marginals, \citet{diakonikolas2010bounded} improved the runtime of \citet{kalai2008agnostically} from $d^{O(\alpha^{-4})}$ to $d^{O(\alpha^{-2})}$ by establishing sharper polynomial approximation bounds.
This is essentially optimal: statistical-query algorithms require time $d^{\Omega(\alpha^{-2})}$~\citep{GoelSQlowerbound2020,diakonikolas2020nearsq,Diakonikolasoptimal2020}, and a nearly matching lower bound holds for all polynomial-time algorithms under the subexponential hardness of Learning with Errors \citep{diakonikolas2023hardness}.
However, this approach is improper, as the resulting classifier need not be a halfspace. 
\citet{diakonikolas2021proper} addressed this limitation by using polynomial regression to identify a subspace containing the normal vector of a near-optimal halfspace; this yields a proper learner with error $\OPT+\alpha$ and runtime $d^{\operatorname{poly}(1/\alpha)}$.

\paragraph{Semi-agnostic learning.}
Motivated by these computational barriers, a related line of work improves computational efficiency by relaxing the agnostic error guarantee.
For homogeneous halfspaces, \citet{awasthi2017localization} developed a
localization-based algorithm achieving $O(\OPT)+\alpha$ error under
isotropic log-concave marginals. Combining localization with polynomial
regression, \citet{daniely2015ptas} obtains an improper approximation scheme
under the uniform spherical marginal, achieving error $(1+\gamma)\OPT+\alpha$
in time polynomial in $d$ and $1/\alpha$ for every fixed $\gamma>0$;
\citet{diakonikolas2021proper} give a proper analogue under Gaussian marginals.
\citet{diakonikolas2020nonconvex} showed that non-convex SGD achieves
$O(\OPT)+\alpha$ error for homogeneous halfspaces under a broad family of
marginals, including isotropic log-concave distributions.
For general halfspaces under Gaussian marginals, \citet{diakonikolas2018nasty}
obtained constant-factor guarantees, even in a stronger contamination model
allowing corruption of both examples and labels, and \citet{DiakonikolasKTN22}
developed the sample-efficient online gradient descent algorithm on which our
private semi-agnostic learner builds.
Complementary work studies convex surrogates for homogeneous halfspaces.
Under well-behaved marginals, including isotropic log-concave distributions,
\citet{frei2021softmargins} obtain $\widetilde{O}(\sqrt{\OPT})+\alpha$ error
using gradient methods on convex losses, while \citet{ji2022logistic} show that
logistic regression alone cannot improve on this and obtain
$\widetilde{O}(\OPT)+\alpha$ by adding a hinge-loss refinement step.

\paragraph{Private learning.}
Distribution-free private learning of halfspaces over infinite domains is
impossible: already thresholds over $\mathbb{R}$ are not privately learnable
\citep{bun2015thresholds}, and private learnability is
characterized by finite Littlestone dimension \citep{littlestone1988learning,alon2019littlestone,bun2020equivalence}.
Over finite domains $X\subset\mathbb{R}^d$, \citet{beimal2019} (improved by
\citet{kaplan2020halfspaces}) give private agnostic learners whose sample complexity
grows with $\log^*|X|$. Generic private realizable-to-agnostic transformations
\citep{alon2020,Li2025} are statistically efficient but not
computationally efficient in general. Efficient private PAC learners exist under margin
assumptions \citep{le2020efficient,bassily2022differentially}, with guarantees
stated for margin-based losses rather than $\OPT$ for the $0$-$1$ loss.

\section{Auxiliary Results}
\label{app:auxiliary-results}
\subsection{Concentration Inequalities}
\label{subsec:concentration-inequalities}

\begin{lemma}[Vector Bernstein Inequality]
\label{lem:vector-bernstein}
Let $U_1, \ldots, U_n$ be independent, mean-zero random vectors in $\mathbb{R}^d$. Suppose that $\|U_i\|_2 \le B$ almost surely for all $i \in \{1, \ldots, n\}$, and let $\sum_{i=1}^n \mathbb{E}\|U_i\|_2^2 \le v$. Then, for every $\beta \in (0, 1)$, with probability at least $1 - \beta$,
\[
  \left\| \frac{1}{n} \sum_{i=1}^n U_i \right\|_2 
  \le \frac{\sqrt{2v \log(2/\beta)}}{n} + \frac{2B \log(2/\beta)}{3n}.
\]
\end{lemma}

\subsection{Differential Privacy Tools}
\label{subsec:privacy-tools}

\begin{definition}[{Differential privacy; \citealp{dwork2006}}]
\label{def:dp}
Let $\epsDP>0$ and $\deltaDP\in[0,1)$. A randomized algorithm
$\mathcal A:\mathcal Z^n\to\mathcal Y$ is
\emph{$(\epsDP,\deltaDP)$-differentially private} if, for
every pair of datasets $S,S'\in\mathcal Z^n$ differing in a single example and every
measurable set $E\subseteq\mathcal Y$,
\[
  \Pr[\mathcal A(S)\in E]\le e^{\epsDP}\,\Pr[\mathcal A(S')\in E]
  +\deltaDP.
\]
\end{definition}

\begin{lemma}[{Post-Processing of DP; \citealp{dwork2014algorithmic}}]\label{lem:post-processing}
    If $M:\cX^n \rightarrow \cY$ is
    $(\epsDP,\deltaDP)$-DP, and $f:\cY \rightarrow \cZ$
    is any randomized function, then the algorithm
    $f \circ M$ is $(\epsDP,\deltaDP)$-DP.
\end{lemma}

\begin{lemma}[{Private convex ERM;
\citealp[Theorem~1.1]{kulkarni2021private}}]
\label{lem:private-lipschitz-erm}
Let \(\Omega\subseteq\mathbb R^D\) be a public, nonempty,
compact convex set of Euclidean diameter \(R\), and let
\(\ell(\cdot;z):\mathbb R^D\to\mathbb R\) be a public convex
\(T\)-Lipschitz loss for every \(z\in\mathcal Z\).
For \(n\ge2\),  \(0<\epsDP\le1/2\), and \(0<\deltaDP\le1/2\),
there exists an \((\epsDP,\deltaDP)\)-DP algorithm under
replacement adjacency such that, for every
\(S=(Z_1,\ldots,Z_n)\in\mathcal Z^n\), its output
\(\widehat v\in\Omega\) satisfies
\[
\mathbb E\!\left[\widehat L_S(\widehat v)\right]
-\min_{v\in\Omega}\widehat L_S(v)
\le
\bigO{
\frac{TR\sqrt{D\log(1/\deltaDP)}}{n\epsDP}
},
\]
where
\(\widehat L_S(v)\coloneqq\frac1n\sum_{i=1}^n\ell(v;Z_i)\),
and the expectation is over the algorithm's randomness.
The algorithm performs \(\bigO{n^2}\) subgradient evaluations
and Euclidean projections onto \(\Omega\).
\end{lemma}

\begin{proof}
Compared to \citet[Theorem~1.1]{kulkarni2021private}, we use \(R\) as an upper bound on the initialization distance and assume convexity and \(T\)-Lipschitzness on all of \(\mathbb R^D\), rather than only on the neighborhood required in the original statement. 
Their complexity guarantee implies \(\bigO{n^2}\) subgradient evaluations in the stated parameter range, with at most as many Euclidean projections, since each iteration uses one projection and at least one subgradient evaluation.
\end{proof}

\begin{lemma}[Validation via the exponential mechanism]
\label{lem:private-finite-validation}
Let \(K,m\in\mathbb N\), and let \(h_1,\ldots,h_K\) be classifiers independent of \(S_{\mathrm{val}}\sim \jointdist^m\).
For \(\epsDP>0\), select \(\widehat j\in[K]\) with probability proportional to \(\exp(-\epsDP m \herr[S_{\mathrm{val}}]{h_j}/2)\).
Conditional on the candidates, selection is \(\epsDP\)-differentially private under replacement.
For \(0<\eta<1\), with probability at least \(1-\eta\),
\[
\err[\jointdist]{h_{\widehat j}}
\le\min_{j\in[K]}\err[\jointdist]{h_j}
+2\sqrt{\frac{\log(4K/\eta)}{2m}}
+\frac{2\log(2K/\eta)}{\epsDP m}.
\]
\end{lemma}

\begin{proof}
Condition on the candidates.
Hoeffding's inequality~\citep{hoeffding1963probability} and a union bound give the first slack term, accounting for the deviations of validation errors from population errors for both the selected classifier and the best candidate.
The score $-m\herr[S_{\mathrm{val}}]{h_j}$ has sensitivity one, so the exponential mechanism ensures privacy and gives the second slack term, accounting for suboptimal selection based on empirical errors~\citep{mcsherry2007mechanism}.
Allocate failure probability $\eta/2$ to each bound.
\end{proof}

\subsection{\texorpdfstring{$\Lone$}{L1} Regression}

\begin{fact}[Random thresholding; {\citealp[Proof of Theorem~5]{kalai2008agnostically}}]
\label{lem:random-threshold}
Let \(\tau\sim\operatorname{Unif}[-1,1]\).
For every \(z\in\reals\) and \(y\in\{-1,1\}\), we have \(\Pr_\tau\bs{\sgn{z-\tau}\ne y} \le \frac12\abs{z-y}.\)
\end{fact}

\begin{lemma}[Uniform deviation for $\Lone$ regression]
\label{lem:expected-regression-deviation}
Let \(\varphi:\mathcal X\to\mathbb R^D\) be a fixed feature map such that
\(\bE\|\varphi(X)\|_2^2\le D\).
Let $L_\jointdist(v)=\bE\,|\ip{v}{\varphi(X)}-Y|$, and let $\widehat L_S(v)$ be its
empirical counterpart on $S\sim \jointdist^n$. Then, for every $R>0$,
\[
\bE_{S\sim \jointdist^n}\sup_{\norm{v}_2\le R}\br{L_\jointdist(v)-\widehat L_S(v)}\le2R\sqrt{\frac Dn}.
\]
\end{lemma}

\begin{proof}
Let $\sigma_1,\dots,\sigma_n$ be i.i.d.\ Rademacher random variables, independent of $S$.
Symmetrization \citep[Lemma~26.2]{shalev2014understanding} and contraction for the
$1$-Lipschitz maps $t\mapsto|t-Y_i|$ \citep[Lemma~26.9]{shalev2014understanding} give
\[
\bE_S\sup_{\norm{v}_2\le R}\br{L_\jointdist(v)-\widehat L_S(v)}
\le\frac{2R}{n}\,\bE_{S,\sigma}\norm{\sum_{i=1}^n\sigma_i\varphi(X_i)}_2.
\]
By Jensen's inequality, and since the cross terms vanish in expectation, the
right-hand side is at most $\frac{2R}{n}\bigl(\sum_{i=1}^n\bE\norm{\varphi(X_i)}_2^2\bigr)^{1/2}
\le2R\sqrt{D/n}$.
\end{proof}

\subsection{Gaussian Marginals}
\label{subsec:gaussian-tools}

For a probability measure $\mu$ on $\mathbb{R}^d$ and $r \in [1,\infty)$, we write \(\|f\|_{L^r(\mu)} \coloneqq \left(\int |f(x)|^r\,d\mu(x)\right)^{1/r}\) and \(
    \|f\|_{L^\infty(\mu)}
    \coloneqq \inf\bigl\{M \ge 0 :
    \mu(\{x : |f(x)| > M\}) = 0\bigr\}.
\)
The inner product on $L^2(\mu)$ is $\langle f,g\rangle_{L^2(\mu)} \coloneqq \int f(x)g(x)\,d\mu(x)$.

\begin{lemma}[{Hermite basis; \citealp[Proposition~11.33]{ODonnell_2014}}]
\label{lem:hermite-basis}
Let $\{h_m\}_{m\ge0}$ be the normalized univariate Hermite polynomials, defined by
\[
e^{tx-t^2/2}
=
\sum_{m=0}^\infty h_m(x)\frac{t^m}{\sqrt{m!}}.
\]
For each multi-index $\kappa=(\kappa_1,\ldots,\kappa_d)\in\bN_0^d$, define $h_\kappa(x)=\prod_{j=1}^d h_{\kappa_j}(x_j)$ and $|\kappa|=\sum_{j=1}^d\kappa_j$.
Then $\{h_\kappa\}_{\kappa\in\bN_0^d}$ is a complete orthonormal basis of $\Ltwo[\cN(0,I_d)]$.
In particular, $\{h_\kappa:|\kappa|\le p\}$ is an orthonormal basis for the polynomials of total degree at most $p$, with dimension $\binom{d+p}{p}$.
\end{lemma}

\begin{lemma}[{Mehler's formula; \citealp[Eq.~6.1.13]{andrews1999special}}]
\label{lem:mehler}
Let $\{h_\kappa\}_{\kappa\in\bN_0^d}$ be the multivariate Hermite polynomials, orthonormal in $\Ltwo[\cN(0,I_d)]$.
For each multi-index $\kappa\in\bN_0^d$, write $|\kappa|=\sum_{j=1}^d\kappa_j$.
For every $x,y\in\reals^d$ and $\rho\in(-1,1)$,
\[
\sum_{\kappa\in\bN_0^d}
\rho^{|\kappa|}h_\kappa(x)h_\kappa(y)
=
\frac{1}{(1-\rho^2)^{d/2}}
\exp\br{
\frac{2\rho\ip{x}{y}
-\rho^2\br{\norm{x}_2^2+\norm{y}_2^2}}
{2(1-\rho^2)}
}.
\]
\end{lemma}

\begin{lemma}[Unit-norm polynomial approximation of halfspaces under Gaussian measure]
\label{lem:unit-norm-approx-gauss}
Let \(d\ge 1\), and let \(\{h_\kappa\}_{\kappa\in\mathbb N_0^d}\) be the orthonormal Hermite basis of
\(\Ltwo[\cN(0,I_d)]\). There exists a universal constant \(C>0\) such that for every \(\alpha\in(0,1)\)
and every halfspace \(c:\mathbb R^d\to\{-1,1\}\), there is a polynomial
\(q(x)=\sum_{|\kappa|\le p}w_\kappa h_\kappa(x)\) of degree \(p\le C\alpha^{-2}\) with
\(\|w\|_2\le 1\) and \(\Lonenorm[\cN(0,I_d)]{c-q}\le\alpha\).
\end{lemma}

\begin{proof}
Write \(c(x)=f_\theta(\langle u,x\rangle)\), where \(f_\theta(z)=\operatorname{sgn}(z-\theta)\) and \(\|u\|_2=1\). 
Let \(B:\mathbb R\to[0,\infty)\) be the one-dimensional smoothing kernel of \citet[Section~4]{diakonikolas2010bounded}, and define \(B_\lambda(t)=\lambda B(\lambda t)\).

By \citet[Section 4.2]{pesenti2026agnostic}, for every
\(\theta\in\mathbb R\),
there exist $\lambda>0$ and an integer
$p\le C\alpha^{-2}$, with $C$ independent of $\theta$, such that \(g_\theta=B_\lambda*f_\theta\) satisfies
\[\Lonenorm[\cN(0,1)]{f_\theta-\Pi_p g_\theta}\le\alpha,\]
where \(\Pi_p(\cdot)\) is the \(\Ltwo[\cN(0,1)]\)-orthogonal projection
onto polynomials of degree at most \(p\).

Since \(\int_{\mathbb R}B_\lambda(t)\,dt=1\)~\citep[Lemma~4.4]{diakonikolas2010bounded}, we have \[|g_\theta(z)| \le\int_{\mathbb R}B_\lambda(t)|f_\theta(z-t)|\,dt =1 \quad\forall z.\]
Hence
\(\Ltwonorm[\cN(0,1)]{\Pi_p g_\theta} \le\Ltwonorm[\cN(0,1)]{g_\theta} \le1.\)

Set \(q(x)=\Pi_p g_\theta(\langle u,x\rangle)\). 
Its degree is at most \(p\), so \(q=\sum_{|\kappa|\le p}w_\kappa h_\kappa\) for some coefficient vector \(w\)~(\Cref{lem:hermite-basis}).
Since \(\|u\|_2=1\), we have \(\langle u,X\rangle\sim\cN(0,1)\) for \(X\sim\cN(0,I_d)\). 
Thus, the approximation error and \(\Ltwo\) norm of \(q\) under \(\mathcal N(0,I_d)\) equal those of \(\Pi_p g_\theta\) under \(\mathcal N(0,1)\).
Finally, orthonormality yields \(\|w\|_2=\Ltwonorm[\cN(0,I_d)]{q}=\Ltwonorm[\cN(0,1)]{\Pi_p g_\theta}\le 1\).
\end{proof}

\begin{fact}[Euclidean covers; {\citealp[Fact~21]{diakonikolas2021proper}}]
\label{fact:euclidean-cover}
For every $k\ge1$ and $\gamma\in(0,1)$, the unit sphere in
$\mathbb R^k$ has an explicit $\gamma$-cover in $\ell_2$ of size at
most $(C/\gamma)^k$, where $C>0$ is a universal constant.
\end{fact}

\begin{fact}[Gaussian halfspace discretization; {\citealp[Fact~22]{diakonikolas2021proper}}]
\label{fact:gaussian-halfspace-discretization}
Let $X\sim N(0,I_d)$, let $u,v\in\mathbb R^d$ be unit vectors, and let
$t_1,t_2\in\mathbb R$. Then
$\mathbb E[|h_{u,t_1}(X)-h_{v,t_1}(X)|]=\bigO{\|u-v\|_2}$ and
$\mathbb E[|h_{u,t_1}(X)-h_{u,t_2}(X)|]=\bigO{|t_1-t_2|}$. Moreover, if
$|t|\ge C\sqrt{\log(1/\gamma)}$, then
$\mathbb E[|h_{v,t}(X)-\operatorname{sign}(t)|]=\bigO{\gamma}$.
\end{fact}

\subsection{Log-Concave Marginals}
\label{subsec:log-concave-tools}

A probability measure is log-concave if its density is of the form $e^{-V}$, where $V$ is convex. 
Log-concavity is preserved under linear projections, and log-concave distributions have finite moments of all orders. 
We use the following quantitative density bound for standardized one-dimensional log-concave distributions.

\begin{fact}[{Bound on the density of log-concave distributions; \citealp[Lemma~6]{kalai2008agnostically}}]\label{fact:lc-density}
Let~$\nu$ be a log-concave distribution on $\bR$ with mean zero and variance one.
Then its density $g$ satisfies $g(x)\le e\cdot e^{-|x|/16}$ for all $x\in\bR$.
\end{fact}

The next two results concern weighted polynomial approximation on $\bR$ with the exponential weight function.
They are not specific to log-concave measures. 
We use them, together with \Cref{fact:lc-density}, in the proof of \Cref{lem:bounded-norm-approx-log-concave}.
Let $\Lpnorm{k}{\cdot}$ denote the $\Lp{k}$ norm with respect to the Lebesgue measure, and let $W(x)\coloneqq e^{-|x|}$.

\begin{lemma}[{Weighted Nikolskii inequality; \citealp[\S9]{lubinsky2007survey}}]
\label{lem:nikolskii}
There is a universal constant $C$ such that, for every $p\ge 1$ and every univariate
polynomial $q$ of degree at most $p$,
\[
\Linfnorm{qW}\le C\log(p+1)\,\Lonenorm{qW}.
\]
\end{lemma}

\begin{lemma}[{Weighted Jackson estimate; \citealp{freud1978approximation}, see also \citealp[\S5]{lubinsky2007survey}}]
\label{lem:weighted-threshold-approx}
Let~$f_\theta(z)=\sgn{z-\theta}$.
There exists a universal constant $C>0$ such that, for every $\theta\in\reals$ and integer $p\ge2$, there is a polynomial $q$ of degree at most $p$ satisfying
\[
    \Lonenorm{(q-f_\theta)W}\le \frac{C}{\log(p+1)}.
\]
\end{lemma}

\begin{proof}
Set $H_\theta=f_\theta W$. 
By \citet[Theorem~2.2]{freud1978approximation}, taking $\delta=1/2$, there exists a polynomial $q$ of degree at most $p$ satisfying
\[
    \Lonenorm{(q-f_\theta)W}
    \le C\Bigg[
    \sup_{|h|\le1/\log p}
      \Lonenorm{H_\theta(\cdot+h)-H_\theta} + 
      \frac{\Lonenorm{H_\theta}}{\log p}
      +\int_{|z|\ge\sqrt p}|H_\theta(z)|\,dz
    \Bigg].
\]

To bound the first term on the right hand side, first observe that $\Lonenorm{W'}=2$. 
Using the fundamental theorem of calculus, we obtain
\(
    \Lonenorm{W(\cdot+h)-W}
    \le |h|\Lonenorm{W'}=2|h|.
\)
Next, $f_\theta(z+h)$ and $f_\theta(z)$ differ only when $z$ lies between $\theta-h$ and $\theta$. 
On this interval, their difference has absolute value $2$. Since $W\le1$,
\(
    \Lonenorm{(f_\theta(\cdot+h)-f_\theta)W} \le 2|h|.
\)
Finally, adding and subtracting $f_\theta(z+h)W(z)$ and applying the triangle inequality gives
\[
    \Lonenorm{H_\theta(\cdot+h)-H_\theta}
    \le \Lonenorm{W(\cdot+h)-W}
       +\Lonenorm{(f_\theta(\cdot+h)-f_\theta)W}
    \le 4|h|.
\]

For the remaining terms, $|H_\theta|=W$ gives $\Lonenorm{H_\theta}=2$ and
\(
    \int_{|z|\ge\sqrt p}|H_\theta(z)|\,dz=2e^{-\sqrt p}.
\)
Substituting these estimates yields $ \Lonenorm{(q-f_\theta)W}
    \le C\left(6/\log p+2e^{-\sqrt p}\right)
    \le C/\log(p+1).$
All constants are independent of $p$ and $\theta$.
\end{proof}

\section{Proofs for Improper Learning with Gaussian Marginals}
\label{app:improper-gaussian}
In this section, we prove~\Cref{thm:improper-agnostic-gaussian}.
We first establish the privacy guarantee of~\Cref{alg:private-polynomial-regression}.

\begin{proposition}\label{prop:improper-gaussian-privacy}
  \Cref{alg:private-polynomial-regression} satisfies
  $(\epsDP,\deltaDP)$-DP under replacement
  of one labeled example.
\end{proposition}

\begin{proof}
  The feature map, the partition into blocks, and the algorithm parameters are data-independent.
  For each example $(x,y)$, the loss $v\mapsto|\langle v,\bar\psi_{p,T}(x)\rangle-y|$ is convex and $T$-Lipschitz, since
  $\|\bar\psi_{p,T}(x)\|_2\le T$.
  As the optimization domain is the public unit ball,~\Cref{lem:private-lipschitz-erm} implies that each regression run is $(\epsDP,\deltaDP)$-DP with respect to its training block.

  All blocks are disjoint, and running the DP gradient descent algorithm with the same privacy parameters independently on each block guarantees $(\epsDP,\deltaDP)$-DP by the guarantees of~\Cref{lem:private-lipschitz-erm} and parallel composition. Our output space being defined for the remainder of the computations via $\hat{v}_j$'s does not violate DP because of post-processing (\Cref{lem:post-processing}). Finally, applying the exponential mechanism (\Cref{lem:private-finite-validation}) on a disjoint validation block of data satisfies $\epsDP$-DP. Therefore, by parallel composition, we obtain $(\epsDP,\deltaDP)$-DP.
\end{proof}

Next, we establish a tail bound for the Hermite feature norm, which controls the bias introduced by clipping. 

\begin{lemma}[Hermite feature norm tail bound]
\label{lem:hermite-clipping}
Let $d,p\ge1$ be integers and $D_p=\binom{d+p}{p}$.
Let $h_1,\ldots,h_{D_p}$ enumerate the multivariate Hermite polynomials of degree at most $p$, orthonormal in~$\Ltwo[\cN(0,I_d)]$.
Define $\psi_p(x)=(h_1(x),\ldots,h_{D_p}(x))$.
For every $\alpha\in(0,1)$ satisfying $d\ge 2p^2/\alpha$ and $X\sim\cN(0,I_d)$,
\[
\bP\bs{\norm{\psi_p(X)}_2 > 2\sqrt{D_p}} \le \alpha.
\]
\end{lemma}

\begin{proof}
Write $V=\norm{\psi_p(X)}_2^2/D_p$.
Orthonormality gives $\bE[V]=1$.
We claim that
\begin{equation}
\label{eq:hermite-variance}
\Var V\le e^{2p^2/d}-1.
\end{equation}
Assuming this claim, Chebyshev's inequality gives
\begin{align*}
\bP\bs{\norm{\psi_p(X)}_2>2\sqrt{D_p}} 
&= \bP\bs{V-1>3} \notag \\
&\le \frac{\bE\bs{(V-1)^2}}{9} \notag \\
&\le \frac{e^{2p^2/d}-1}{9} \notag \\
&\le \frac{e^\alpha-1}{9} && (\text{since } d \ge 2p^2/\alpha)\\
&\le \frac{(e-1)\alpha}{9} && (\text{by convexity of } e^x \text{ for } \alpha \in [0,1]) \notag \\
&\le \alpha. \notag
\end{align*}

It remains to establish \eqref{eq:hermite-variance}.
For every integer $m\ge0$, define $K_m(x)=\sum_{|\kappa|\le m}h_\kappa(x)^2.$
In particular, $K_p(X)=D_pV$.
To compute the second moment of $K_p(X)$, we use
the generating function $F(z,x)=\sum_{m=0}^\infty K_m(x)z^m$.
Mehler's formula~(restated in \Cref{lem:mehler}) gives
\[
F(z,x)
=\frac1{1-z}
\sum_{\kappa\in\bN_0^d}z^{|\kappa|}h_\kappa(x)^2
=\frac{\exp\br{\frac{z}{1+z}\norm{x}_2^2}}
{(1-z)(1-z^2)^{d/2}}.
\]
Since $\norm{X}_2^2\sim\chi_d^2$, its moment-generating function gives, for sufficiently small $z,w>0$,
\[
\begin{aligned}
\bE[F(z,X)F(w,X)]
=\frac1{(1-z)^{d+1}(1-w)^{d+1}} \br{1-\frac{4zw}{(1-z)(1-w)}}^{-d/2}.
\end{aligned}
\]
Expanding the last factor using the binomial series
and comparing coefficients of $z^pw^p$ yields
\[
\bE[K_p(X)^2]
=
\sum_{k=0}^p
\frac{4^k(d/2)_k}{k!}
\binom{d+p}{p-k}^2,
\]
where $(a)_k=\prod_{j=0}^{k-1}(a+j)$, with $(a)_0=1$.
We therefore arrive at
\[
\begin{aligned}
\bE[V^2]
&=\sum_{k=0}^p
\frac{4^k(d/2)_k}{k!}
\br{\frac{\binom{d+p}{p-k}}{D_p}}^2\\
&\le\sum_{k=0}^p\frac{(2p^2/d)^k}{k!}\\
&\le e^{2p^2/d},
\end{aligned}
\]
where the first inequality follows from
\(
\binom{d+p}{p-k}/D_p
=\prod_{j=0}^{k-1}\frac{p-j}{d+1+j},
\)
together with $p-j\le p$ and
$(d/2+j)/(d+1+j)^2\le 1/(2d)$.
Since $\bE[V]=1$, \eqref{eq:hermite-variance} follows.
\end{proof}

We now turn to the proof of \Cref{thm:improper-agnostic-gaussian}.
We use~\Cref{alg:private-polynomial-regression} with the degree-$p$ Hermite feature map $\psi_p$. 
Fix a universal constant $c\in(0,1/4]$ and a sufficiently large universal constant $C$, depending only on $c$.
Set
\begin{subequations}\label{eq:alg-parameters}
\begin{equation}\label{eq:gaussian-parameters}
p=\left\lceil C\alpha^{-2}\right\rceil,
\quad
D_p=\binom{d+p}{p},
\quad
T=2\sqrt{D_p}, \quad
k=\left\lceil\log_2\frac{2}{\beta}\right\rceil,
\quad
M=\left\lceil\frac{C}{\alpha}\right\rceil.
\end{equation}
Since \(p\le(C+1)\alpha^{-2}\), choosing
\(C_{\mathrm{dim}}\ge 2(C+1)^2/c\) ensures that
\(d\ge C_{\mathrm{dim}}\alpha^{-5}\) implies
\(d\ge2p^2/(c\alpha)\).
Thus~\Cref{lem:hermite-clipping}  applies at level \(c\alpha\).
Let $n_0\ge 2$ be the smallest integer satisfying
\begin{equation}\label{eq:gaussian-training-size}
C\left(
\frac{T\sqrt{D_p\log(1/\deltaDP)}}{\epsDP n_0}
+
\sqrt{\frac{D_p}{n_0}}
\right)
\le c\alpha,
\end{equation}
and set
\begin{equation}\label{eq:gaussian-validation-size}
m=\left\lceil
  C\left(\frac{1}{\alpha^2}
    +\frac{1}{\epsDP\alpha}\right)
  \log\frac{kM}{\beta}
\right\rceil,
\qquad
n=kn_0+m.
\end{equation}
\end{subequations}

\begin{proof}[Proof of \Cref{thm:improper-agnostic-gaussian}]
\Cref{prop:improper-gaussian-privacy} establishes privacy.
It remains to establish the accuracy, sample complexity, and running time bounds.
Write \(\xmarginal=\mathcal N(0,I_d)\), and recall the population regression loss
$
L_\jointdist(v)
=\bE_{(X,Y)\sim \jointdist}
\left|\ip{v}{\bar\psi_{p,T}(X)}-Y\right|$, and its minimum $L^\star \coloneqq \min_{\norm{v}_2\le1}L_\jointdist(v)$.
Additionally, for the training block \(S_j\), write
\(
\widehat L_{S_j}(v)
\coloneqq
n_0^{-1}\sum_{(X_i,Y_i)\in S_j}
\left|\ip{v}{\bar\psi_{p,T}(X_i)}-Y_i\right|
\)
for the empirical regression loss.

We first bound the expected excess regression loss of a
single training run.
Orthonormality and clipping give
\(\bE_{\xmarginal}\|\bar\psi_{p,T}(X)\|_2^2\le D_p\) and
\(\|\bar\psi_{p,T}(x)\|_2\le T\).
The loss is convex and \(T\)-Lipschitz on all of \(\mathbb R^{D_p}\), and the optimization domain is the public unit ball, of diameter \(2\).
Fix a population minimizer \(v^\star\in\argmin_{\norm{v}_2\le1}L_\jointdist(v)\), so that \(L_\jointdist(v^\star)=L^\star\).
Adding and subtracting \(\widehat L_{S_j}(\widehat v_j)\) and \(\widehat L_{S_j}(v^\star)\) give
\begin{equation}
\label{eq:improper-regression-decomposition}
\begin{aligned}
L_\jointdist(\widehat v_j)-L^\star
&=
\bigl[
L_\jointdist(\widehat v_j)
-\widehat L_{S_j}(\widehat v_j)
\bigr]
+
\bigl[
\widehat L_{S_j}(\widehat v_j)
-\widehat L_{S_j}(v^\star)
\bigr]+
\bigl[
\widehat L_{S_j}(v^\star)-L^\star
\bigr]\\
&\le
\sup_{\norm{v}_2\le1}
\bigl(L_\jointdist(v)-\widehat L_{S_j}(v)\bigr)
+
\Bigl[
\widehat L_{S_j}(\widehat v_j)
-\min_{\norm{v}_2\le1}\widehat L_{S_j}(v)
\Bigr]+
\bigl[
\widehat L_{S_j}(v^\star)-L^\star
\bigr].
\end{aligned}
\end{equation}
The inequality uses \(\norm{\widehat v_j}_2\le1\).
Since \(v^\star\) is fixed and the examples in \(S_j\) are sampled from \(\jointdist\),
\(
\bE_{S_j}\widehat L_{S_j}(v^\star)
=L_\jointdist(v^\star)=L^\star.
\)
Thus the final term in \eqref{eq:improper-regression-decomposition} has mean zero.
Taking expectations over $S_j$ and the randomness of the private optimizer, \Cref{lem:private-lipschitz-erm,lem:expected-regression-deviation} therefore give
\begin{equation}\label{eq:gaussian-expected-excess}
\bE[L_\jointdist(\widehat v_j)-L^\star]
\le
\bE\!\left[
\widehat L_{S_j}(\widehat v_j)
-\min_{\norm{v}_2\le1}\widehat L_{S_j}(v)
\right]
+2\sqrt{\frac{D_p}{n_0}}
\le c\alpha,
\end{equation}
where the last inequality follows from the choice of $n_0$ in \eqref{eq:gaussian-training-size}.

We then use the standard confidence-amplification argument based on
independent repetition; see, e.g.,
\citet[proof of Theorem~26]{shalev2010learnability}.
By \eqref{eq:gaussian-expected-excess} and Markov's inequality,
$
\bP\!\left(
L_\jointdist(\widehat v_j)-L^\star>2c\alpha
\right)
\le 1/2.
$
Since the $k$ training runs are independent,
\(
\bP(
\min_{j\in[k]} L_\jointdist(\widehat v_j)>L^\star+2c\alpha
)
\le 2^{-k}
\le \beta/2.
\)
Thus, with probability at least $1-\beta/2$,
\begin{equation}\label{eq:gaussian-training-success}
\min_{j\in[k]} L_\jointdist(\widehat v_j)
\le L^\star+2c\alpha.
\end{equation}

Next, we argue about the correctness of the choice of our threshold, which requires our grid over $[-1,1]$ to be fine enough that the bias due to discretization is small, but not so fine that we blow up in terms of our sample complexity and running time when applying the exponential mechanism.
If we select a threshold uniformly at random from the continuous interval $[-1,1]$, then \Cref{lem:random-threshold} bounds the expected classification error under a uniform threshold $\tau\sim\operatorname{Unif}[-1,1]$ by half the absolute loss. Then the expected $0$-$1$ loss preserves our agnostic learning error requirements.
However, rounding $\tau$ to the midpoint of its grid interval yields a threshold that is uniform over $\{t_1,\ldots,t_M\}$.
At a fixed $x$, rounding changes the prediction only if the score
$\langle\widehat v_j,\bar\psi_{p,T}(x)\rangle$ lies between $\tau$ and the midpoint $\tau$ is rounded to, which happens with probability at most $1/(2M)$. Hence,
the expectation of the population error of $\widehat{v}_j$ over the random sampling from the discretized grid is $M^{-1}\sum_{\ell=1}^M\err[\jointdist]{h_{j,\ell}}\le L_\jointdist(\widehat v_j)/2+1/(2M)$.
Since $\min_{\ell}\err[\jointdist]{h_{j,\ell}} \leq M^{-1}\sum_{\ell=1}^M\err[\jointdist]{h_{j,\ell}}$, conditioned on the event in \eqref{eq:gaussian-training-success} and taking the minimum over all $\widehat{v}_j$'s, we have
\begin{equation}\label{eq:gaussian-candidate-error}
\min_{j,\ell}\err[\jointdist]{h_{j,\ell}}
    \le
    \frac12\min_{j\in[k]}L_\jointdist(\widehat v_j)
    +\frac1{2M}
    \le
    \frac12L^\star+2c\alpha.
\end{equation}

Let \(h\) denote the classifier selected by the exponential mechanism on \(S_{\mathrm{val}}\).
Conditional on the training outputs,
\Cref{lem:private-finite-validation}, with $K=kM$,
$\eta=\beta/2$, and $m$ as in \eqref{eq:gaussian-validation-size},
gives $\err[\jointdist]{h} \le \min_{j,\ell}\err[\jointdist]{h_{j,\ell}}+c\alpha$ with probability at least $1-\beta/2$.
Thus, with probability at least \(1-\beta\),
\begin{equation}
\label{eq:improper-classification-bound}
    \err[\jointdist]{h}\le L^\star/2+3c\alpha,
\end{equation}
which comes due to sampling and the noise due to the exponential mechanism.

We next compare $L^\star$ with the error of the best halfspace. 
Fix $c_0\in\mathcal C$. 
By \Cref{lem:unit-norm-approx-gauss}, and the choice of the degree $p$ in \eqref{eq:gaussian-parameters}, there is a polynomial $q=\ip{v_q}{\psi_p}$ with $\norm{v_q}_2\le1$ and $\Lonenorm[\xmarginal]{q-c_0}\le c\alpha$.
Write $s(x)\coloneqq T/\max\{T,\norm{\psi_p(x)}_2\}$, so that $\bar\psi_{p,T}(x)=s(x)\psi_p(x)$.
Since $0<s(x)\le1$ and $|y|=1$,
\begin{align*}
\bigl|\ip{v_q}{\bar\psi_{p,T}(x)}-y\bigr|
&=|s(x)q(x)-y|\\
&\le s(x)|q(x)-y|+1-s(x)\\
&\le |q(x)-c_0(x)|+|c_0(x)-y|+1-s(x).
\end{align*}
Taking expectations over a population example \((X,Y)\sim\jointdist\) yields
\begin{equation}
\label{eq:improper-clipping-comparison}
L_\jointdist(v_q)
\le 2\err[\jointdist]{c_0}
+\Lonenorm[\xmarginal]{q-c_0}
+\bE_\xmarginal[1-s(X)].
\end{equation}
Moreover, since \(0\le1-s(x)\le\mathbf 1\{\|\psi_p(x)\|_2>T\}\),~\Cref{lem:hermite-clipping}, applied at level \(c\alpha\), gives
\(
\mathbb E_\mu[1-s(X)]
\le
\Pr_\mu(\|\psi_p(X)\|_2>T)
\le c\alpha.
\)
Since $\norm{v_q}_2\le1$, we conclude that $L^\star\le L_\jointdist(v_q) \le2\err[\jointdist]{c_0}+2c\alpha$.
Taking the infimum over $c_0\in\mathcal C$ gives
\begin{equation}\label{eq:gaussian-comparator}
L^\star
\le 2\OPT+2c\alpha.
\end{equation}
Combining \eqref{eq:improper-classification-bound}
and \eqref{eq:gaussian-comparator}, we obtain the desired
\[
\err[\jointdist]{h}
\le \OPT+4c\alpha
\le \OPT+\alpha,
\]
where the last inequality uses $c\le1/4$.

The choices in \eqref{eq:gaussian-training-size} and \eqref{eq:gaussian-validation-size} satisfy
\[
n_0
=\softO{
\frac{D_p}{\alpha^2}
+\frac{D_p}{\epsDP\alpha}
},
\qquad
m
=\softO{
\left(
\frac1{\alpha^2}
+\frac1{\epsDP\alpha}
\right)
\log\frac{2}{\beta}
},
\]
where the bound on $n_0$ uses $T\sqrt{D_p}=2D_p$.
Since $k=\bigO{\log(2/\beta)}$ and $n=kn_0+m$, this yields the stated sample bound on $n$.

Finally, we analyse the running time.
Computing the clipped Hermite features requires $\bigO{pD_p}$ arithmetic operations per example.
Given these features, subgradient evaluation and projection onto the Euclidean unit ball each take $\bigO{D_p}$ time, so \Cref{lem:private-lipschitz-erm} gives a total cost of $\bigO{kn_0^2D_p}$ for the $k$ private ERM runs.
Evaluating the $kM$ candidate classifiers on the validation sample requires $\bigO{kMmD_p}$ operations.
Thus, the overall running time is
\[
\bigO{npD_p+kn_0^2D_p+kMmD_p}
=\poly{n},
\]
since $M\le p<D_p\le n_0$ and $kn_0+m=n$.
\end{proof}

\section{Private Improper Agnostic Learning: Log-Concave Marginals}
\label{app:improper-log-concave}
We extend the learner of~\Cref{sec:improper} to known, full-dimensional\footnote{
  The full-dimensionality assumption is without loss of generality since the marginal is known.
} log-concave marginals.
The regression, confidence amplification, thresholding, and private validation steps are unchanged.
We replace the Hermite features by a polynomial basis that is orthonormal under the marginal.
The approximation guarantee in \Cref{lem:bounded-norm-approx-log-concave} allows us to restrict regression to a coefficient ball of constant radius.
We use the known marginal to choose a clipping threshold $T$ that controls the additional regression loss while keeping the feature norms bounded.

\begin{theorem}[DP improper agnostic learner for log-concave marginals]\label{thm:improper-agnostic-log-concave}
Let $\alpha, \beta \in (0, 1)$ and $\epsDP, \deltaDP \in (0, 1/2]$.
Set $p = \lceil \exp(C/\alpha) \rceil$ for a sufficiently large universal constant $C > 0$,
let $D_p = \binom{d+p}{p}$, and let $T$ be as in~\eqref{eq:log-concave-threshold}.
Let $P$ be any distribution on $\mathbb{R}^d \times \{\pm 1\}$ with a known,
full-dimensional log-concave marginal $\mu$. Then the modification of
  \Cref{alg:private-polynomial-regression} described in~\Cref{subsec:logconcave-algorithm} is $(\varepsilon_{\mathrm{DP}}, \delta_{\mathrm{DP}})$-DP
and, given
\[
  n = \softO{\frac{D_p}{\alpha^2} + \frac{T\sqrt{D_p}}{\epsDP\,\alpha}}
\]
i.i.d.\ samples from $P$, returns a classifier $h$ satisfying
$\err[\jointdist]{h} \le \OPT + \alpha$ with probability at least
$1 - \beta$. The algorithm runs in time $\operatorname{poly}(n)$, provided that constructing
the orthonormal feature map, evaluating it at an input, and computing $T$ from public
information each take time $\operatorname{poly}(D_p)$.
\end{theorem}

\noindent
As we show in~\Cref{remark:all-dimensions}, $T=\sqrt{D_p/(c\alpha)}$ is an admissible clipping threshold.
Consequently, the sample bound is at most $\softO{D_p/\alpha^2+ D_p/(\epsDP\alpha^{3/2})}$.

\subsection{The Algorithm}
\label{subsec:logconcave-algorithm}

Let $\xmarginal$ be a known, full-dimensional log-concave distribution on $\bR^d$. We modify~\Cref{alg:private-polynomial-regression} by replacing the Hermite features with a polynomial basis orthonormal under $\xmarginal$, choosing a clipping threshold adapted to $\xmarginal$, and changing the optimization domain.
We increase the polynomial degree to achieve the required
approximation accuracy.
We keep $k$, $M$, and $m$ as in~\eqref{eq:gaussian-parameters} and~\eqref{eq:gaussian-validation-size}, and replace the block size
$n_0$ by~\eqref{eq:log-concave-block-size}.
The private optimization procedure, independent repetitions, thresholding, and private validation remain unchanged.
We discuss the changes to~\Cref{alg:private-polynomial-regression} in greater detail in the remainder of this section.

First, for the existence of a polynomial with good approximating guarantees, we increase the degree $p=\lceil\exp(C/\alpha)\rceil$ for a sufficiently large universal constant $C>0$.
For the optimization domain, we replace the unit coefficient ball in~\Cref{alg:private-polynomial-regression} with $ B_2^{D_p}(R_0)\coloneqq \{v\in\bR^{D_p}:\|v\|_2\le R_0\}$, where $R_0$ is the universal constant in \Cref{lem:bounded-norm-approx-log-concave}.

Next, we recall how to construct an orthonormal basis for the polynomials of degree at most $p$ in $\Ltwo[\xmarginal]$.
Let $m_p(x)=(x^\kappa)_{|\kappa|\le p}$ be the vector of monomials of degree at most $p$.
Following the standard moment-matrix orthonormalization~(\citealp{Dunkl_Xu_2014}), define the population Gram matrix $G_p=\mathbb E_\xmarginal[m_p(X)m_p(X)^\top]$ and set $\psi_p(x)=G_p^{-1/2}m_p(x)$.
The entries of $G_p$ are the moments of $\xmarginal$ through degree $2p$.
Log-concavity ensures that all moments are finite, while full dimensionality guarantees that $G_p$ is positive definite.
Since $G_p^{-1/2}$ is invertible, the coordinates of $\psi_p$ span the polynomials of degree at most $p$.
Moreover, $\mathbb E_\xmarginal[\psi_p(X)\psi_p(X)^\top]=I_{D_p}$,
so they form an orthonormal basis of this polynomial space in $\Ltwo[\xmarginal]$.
In particular, $\Ltwonorm[\xmarginal]{\langle v,\psi_p\rangle}=\|v\|_2$ and $\mathbb E_\xmarginal\|\psi_p(X)\|_2^2=D_p$.
This construction based on the moment-matrix provides a general way to obtain the required orthonormal basis. When an explicit basis is available, it may be used directly without forming \(G_p\), provided its construction and evaluation satisfy the computational assumption in~\Cref{thm:improper-agnostic-log-concave}. For example, the normalized Hermite basis for the standard Gaussian marginal can be constructed and evaluated in \(\operatorname{poly}(D_p)\) arithmetic operations.

Finally, we discuss clipping.
As in the Gaussian case, we radially clip the feature vectors to obtain a uniform bound on their norms: $\bar\psi_{p,T}(x) \coloneqq {T\psi_p(x)}/\max\{T,\|\psi_p(x)\|_2\}.$
Since $\xmarginal$ is known, we can choose the clipping threshold directly from the marginal.
For a sufficiently small universal constant $c>0$, set
\begin{equation}
  \label{eq:log-concave-threshold}
  T=\min\left\{t>0:
  \mathbb E_\xmarginal\left[
    \left(1-\frac{t}{\|\psi_p(X)\|_2}\right)_+
    \right]\le c\alpha
  \right\},
\end{equation}
where $(x)_+=\max\{x,0\}$.
The argument for~\eqref{eq:universal-clipping} uses only
orthonormality and therefore also applies to the present feature map.
Since
\((1-t/\|\psi_p(x)\|_2)_+
\le\mathbf 1\{\|\psi_p(x)\|_2>t\}\),
the choice \(t=\sqrt{D_p/(c\alpha)}\) is admissible,
and hence \(T\le\sqrt{D_p/(c\alpha)}\).
Depending on the marginal and the parameter regime, marginal-dependent choices may yield a smaller threshold.
For example, if $\xmarginal=N(0,I_d)$ and $d\ge2p^2/(c\alpha)$, \Cref{lem:hermite-clipping} shows that $T\le2\sqrt{D_p}$, so the privacy term improves to $\bigO{D_p/(\epsDP\alpha)}$.\footnote{
  \Cref{lem:hermite-clipping} is stated for Hermite features, but any two orthonormal bases of the degree-$\le p$ polynomials in $\Ltwo[\xmarginal]$ differ by an orthogonal matrix, so $\|\psi_p(x)\|_2$ is the same for both.
}

\subsection{Analysis}

In this section, we prove~\Cref{thm:improper-agnostic-log-concave}.
We first establish privacy.
\begin{proposition}
  \label{prop:improper-log-concave-privacy}
  The learner in~\Cref{thm:improper-agnostic-log-concave} satisfies
  $(\epsDP,\deltaDP)$-DP under replacement
  of one labeled example.
\end{proposition}

\begin{proof}
  The feature map, clipping threshold, coefficient ball, and sample partition are public. 
  Since \(\|\bar\psi_{p,T}(x)\|_2\le T\), the loss is convex and \(T\)-Lipschitz in $v$ on all of \(\mathbb R^{D_p}\).
  Thus, each regression run is $(\epsDP,\deltaDP)$-DP by \Cref{lem:private-lipschitz-erm}.
  The neighboring-dataset argument in the proof of \Cref{prop:improper-gaussian-privacy} then applies unchanged, establishing $(\epsDP,\deltaDP)$-DP for the full learner.
\end{proof}

We next prove the bounded-norm polynomial approximation guarantee.
The key ingredient is the exponential density bound for one-dimensional log-concave distributions (\Cref{fact:lc-density}), which allows us to transfer weighted polynomial approximation bounds to the $\Lone$ and $\Ltwo$ norms under the marginal.
The result extends to arbitrary halfspaces by considering their one-dimensional projections.\looseness=-1

\begin{lemma}[Bounded-norm polynomial approximation of halfspaces under log-concave measures]
  \label{lem:bounded-norm-approx-log-concave}
  There exist universal constants $C,R_0\ge1$ such that the following
  holds. Let $\xmarginal$ be a log-concave probability measure on $\reals^d$. For every integer $p\ge2$ and halfspace
  $c:\reals^d\to\bc{-1,1}$, there is a polynomial $q$ of degree at
  most $p$ satisfying $\Lonenorm[\xmarginal]{q-c}\le C/\log(p+1)$ and
  $\Ltwonorm[\xmarginal]{q}\le R_0$.
  In particular, taking $p=\lceil\exp(C/\alpha)\rceil$ gives
  $\Lonenorm[\xmarginal]{q-c}\le\alpha$.
\end{lemma}
\begin{proof}
  Consider a log-concave probability measure \(\nu\) on \(\reals\) with mean zero and variance one,  and set \(f_{\theta}\br{z} = \sgn{z-\theta}\). We will first prove that there exists a polynomial \(r\) of degree at most \(p\) satisfying \begin{equation}\label{eq:one-dim-bound}
    \Lonenorm[\nu]{r-f_\theta}\le C/\log(p+1) \enskip\text{and}\enskip \Ltwonorm[\nu]{r-f_\theta}^2\le C.
  \end{equation}

  \noindent Assuming the above is true, we next extend it to \(\reals^d\). Write the halfspace \(c\br{x} = \sgn{\ip{u}{x}-\theta}\). For $X\sim\xmarginal$, consider the mean \(a=\bE_\xmarginal\ip{u}{X}\) and variance \(b^2=\operatorname{Var}_\xmarginal\br{\ip{u}{X}}>0\).\footnote{If $b=0$, take $q(x)\equiv\sgn{a-\theta}$.} Let $\nu$ denote the distribution
  of the normalized random variable $Z=(\langle u,X\rangle-a)/b$.
  Then $\nu$ is log-concave, with mean zero and variance one. Apply~\Cref{eq:one-dim-bound} to
  $f_{(\theta-a)/b}$ and set $q(x)=r\br{\br{\langle u,x\rangle-a}/b}$. Since $q(X)=r(Z)$ and $c(X)=f_{(\theta-a)/b}(Z)$, this preserves the degree of the polynomial and yields $\Lonenorm[\xmarginal]{q-c}\le C/\log(p+1)$ and \(\Ltwonorm[\xmarginal]{q - c}^2\le C\). Since
  $\Ltwonorm[\xmarginal]{c}=1$, using the triangle inequality  on \(q = \br{q-c}+c\) gives
  \[
    \Ltwonorm[\xmarginal]{q}
    \le \Ltwonorm[\xmarginal]{q-c}+\Ltwonorm[\xmarginal]{c}
    \le \sqrt C+1\eqqcolon R_0.
  \]

  It remains to prove~\Cref{eq:one-dim-bound}.
  Let \(W\br{z}=\exp\br{-\abs{z}/32}\) Then using~\citet[Lemma 6]{kalai2008agnostically}~(restated in~\Cref{fact:lc-density}), the density satisfies
  \begin{equation}
    \label{eq:log-concave-density-envelope}
    g(z)\le e\exp(-|z|/16).
  \end{equation}
  Note that $f_\theta\br{z}W\br{z}$ is the product of a sign function and an
  exponentially decaying weight, so \(|f_\theta(z)W(z)|=e^{-|z|/32}\). Then,  the approximation theorem of \citet{freud1978approximation}~(restated in~\Cref{lem:weighted-threshold-approx}) gives a polynomial \(r\) of degree at most \(p\) satisfying
  \begin{equation}
    \label{eq:log-concave-weighted-approximation}
    \Lonenorm{(r-f_\theta)W}\le \frac{C}{\log(p+1)}.
  \end{equation}
  We now bound the \(L_2\) error:
  \begin{equation}\label{eq:lemma-c1-l2}
    \begin{aligned}
      \Ltwonorm{(r-f_\theta)W}^2
      \le \Linfnorm{(r-f_\theta)W}\Lonenorm{(r-f_\theta)W}
       & \le \br{\Linfnorm{rW}+ \Linfnorm{f_\theta W}}\frac{C}{\log(p+1)} \\
       & \le C\log(p+1)\,\frac{C}{\log(p+1)}
      \le C.
    \end{aligned}
  \end{equation}
  Here the first inequality follows from H\"older's inequality and the second uses~\Cref{eq:log-concave-weighted-approximation} and triangle inequality. The third inequality follows from \citet[Theorem~3.1]{mhaskar1984extremal}~(restated in~\Cref{lem:nikolskii}) and observing that \(\Linfnorm{f_\theta W}\le1\).
  Now using \(g\le eW^2\) and \(W\le 1\) we can transfer \Cref{eq:log-concave-weighted-approximation,eq:lemma-c1-l2} to a norm with respect to measure induced by \(\nu\), to get~\Cref{eq:one-dim-bound}.
  This completes the proof.
\end{proof}

We now turn to the proof of~\Cref{thm:improper-agnostic-log-concave}.
Fix $c\in(0,1/4]$ as in~\eqref{eq:log-concave-threshold}.
The block-size condition differs from the Gaussian case only through the radius \(R_0\) and the threshold \(T\): let \(n_0\ge2\) be the smallest integer satisfying
\begin{equation}
  \label{eq:log-concave-block-size}
  C\left[
\frac{R_0T\sqrt{D_p\log(1/\deltaDP)}}{\epsDP n_0}
+
R_0\sqrt{\frac{D_p}{n_0}}
\right]
\le c\alpha.
\end{equation}

\begin{proof}[Proof of \Cref{thm:improper-agnostic-log-concave}]
  \Cref{prop:improper-log-concave-privacy} provides the privacy
  guarantee. We now prove the accuracy and running time bounds.
  Write $L_\jointdist(v)=\mathbb E_{(X,Y)\sim \jointdist}
    |\langle v,\bar\psi_{p,T}(X)\rangle-Y|$ and
  $L^\star=\min_{\|v\|_2\le R_0}L_\jointdist(v)$.
  The minimum exists since the domain is compact and the clipped
  loss is continuous in $v$.

  We first compare $L^\star$ with the error of the best halfspace.
  Fix $c_0\in\mathcal C$. By \Cref{lem:bounded-norm-approx-log-concave}, our choice of $p$
  gives a polynomial $q=\langle v_q,\psi_p\rangle$ such that $\Lonenorm[\xmarginal]{q-c_0}\le c\alpha$
  and $\|v_q\|_2=\Ltwonorm[\xmarginal]{q}\le R_0$.
  Let $s(x)=T/\max\{T,\|\psi_p(x)\|_2\}$. Since $1-s(x)=(1-T/\|\psi_p(x)\|_2)_+$,
  \eqref{eq:log-concave-threshold} gives $\bE_\xmarginal[1-s(X)]\le c\alpha$.
  Substituting this into \eqref{eq:improper-clipping-comparison} gives
  $L_\jointdist(v_q)\le2\err[\jointdist]{c_0}+2c\alpha$.
  Since $\|v_q\|_2\le R_0$, taking the infimum over $c_0\in\mathcal C$ yields
  \begin{equation}
    \label{eq:log-concave-comparator-loss}
    L^\star\le2\OPT+2c\alpha.
  \end{equation}

  Next, we bound the expected excess regression loss of a single
  training run.
  The loss is convex and \(T\)-Lipschitz on all of
\(\mathbb R^{D_p}\), the optimization domain has diameter
\(2R_0\), and
\(\mathbb E_\mu\|\bar\psi_{p,T}(X)\|_2^2\le D_p\).
  Applying the decomposition in
  \eqref{eq:improper-regression-decomposition}, we obtain
  \begin{align*}
\mathbb E[L_P(\widehat v_j)-L^\star]
&\le
\mathbb E\!\left[
\widehat L_{S_j}(\widehat v_j)
-
\min_{\|v\|_2\le R_0}\widehat L_{S_j}(v)
\right]
+
\mathbb E\sup_{\|v\|_2\le R_0}
\left(L_P(v)-\widehat L_{S_j}(v)\right)\\
&\le
C\frac{R_0T\sqrt{D_p\log(1/\deltaDP)}}{\epsDP n_0}
+
2R_0\sqrt{\frac{D_p}{n_0}} \le c\alpha.
\end{align*}
  where the second inequality follows from
  \Cref{lem:private-lipschitz-erm,lem:expected-regression-deviation},
  and the last from \eqref{eq:log-concave-block-size}.

  Since the training runs are independent and the thresholding
  and validation steps use the same choices of $k$, $M$, and $m$,
  the argument leading to
  \eqref{eq:improper-classification-bound} applies with
  $L^\star=\min_{\norm{v}_2\le R_0}L_\jointdist(v)$.
  Consequently, using~\eqref{eq:log-concave-comparator-loss}, $\err[\jointdist]{h} \le  \OPT+\alpha$ with probability at least $1-\beta$.

  It remains to bound the sample size and running time.
  Since $R_0$ is a universal constant,
  \eqref{eq:log-concave-block-size} gives
  \[
    n_0
=
O\!\left(
\frac{D_p}{\alpha^2}
+
\frac{T\sqrt{D_p\log(1/\deltaDP)}}{\epsDP\alpha}
\right),
  \]
  independent of $\beta$, and $k=\bigO{\log(2/\beta)}$.
  The validation size $m$ is absorbed by this bound. Indeed, $D_p\ge1$, and every
  point satisfies $\|\psi_p(x)\|_2\ge1$ because the constant polynomial has unit
  $\Ltwo[\xmarginal]$ norm. Hence, the expectation in~\eqref{eq:log-concave-threshold} is at
  least $(1-t)_+$, so $T\ge1-c\alpha\ge3/4$.
  This proves the stated bound on $n=kn_0+m$. Since
  $T\le\sqrt{D_p/(c\alpha)}$ by~\eqref{eq:universal-clipping}, the bound is always at most
  $\softO{D_p/\alpha^2+D_p/(\epsDP\alpha^{3/2})}$.

  Finally, we bound the running time, measured in arithmetic operations. By the computational
  hypothesis in~\Cref{thm:improper-agnostic-log-concave}, computing $G_p^{-1/2}$ and $T$
  takes $\operatorname{poly}(D_p)$ operations. For each example, evaluating
  $\psi_p(x)=G_p^{-1/2}m_p(x)$ takes $\bigO{pD_p}$ operations for the monomials and $\bigO{D_p^2}$
  for the matrix--vector product, and clipping takes $\bigO{D_p}$ more. The remaining steps are
  those of~\Cref{alg:private-polynomial-regression} with the unit ball replaced by
  $B_2^{D_p}(R_0)$. Projection onto this ball still takes $\bigO{D_p}$ operations, so, as in the
  proof of~\Cref{thm:improper-agnostic-gaussian}, the private ERM runs and the validation
  step take $\bigO{kn_0^2D_p}$ and $\bigO{kMmD_p}$ operations, respectively. Since $p<D_p$, the total is
  \[
    \operatorname{poly}(D_p)+\bigO{nD_p^2+kn_0^2D_p+kMmD_p}=\operatorname{poly}(n),
  \]
  where the last equality uses $M\le p<D_p\le n_0$ and $kn_0+m=n$.
\end{proof}

\section{Proofs for Proper Learning with Gaussian Marginals}
\label{app:proper-gaussian}
\begin{proof}[Proof of~\Cref{lem:proper-finite-class}]
  We follow the subspace construction in the proof of
  \citet[Proposition~5]{diakonikolas2021proper}. Write
  $\xmarginal=\cN(0,I_d)$.
  Let $q^*$ be the orthogonal projection of
  $r(x)=\bE[U\mid X=x]$ onto $\cP_p$. The squared-loss identity gives
  $\Ltwonorm[\xmarginal]{q-q^*}\le c_0\alpha$. Since $\abs{r}\le1$, we also have
  $\Ltwonorm[\xmarginal]{q^*}\le1$ and $\Ltwonorm[\xmarginal]{q}\le1+c_0\alpha$.
  Consequently, for every $Q\in\cP_p$,
  \begin{equation}\label{eq:proper-approximate-orthogonality}
    \abs{\bE\bs{(U-q(X))Q(X)}}
    =\abs{\bE\bs{(q^*(X)-q(X))Q(X)}}
    \le c_0\alpha\Ltwonorm[\xmarginal]{Q}.
  \end{equation}

  \noindent Set $M=\bE_\xmarginal\bs{\nabla q(X)\nabla q(X)^\top}$, and let $V$ be
  the span of its eigenvectors with eigenvalues at least
  $\alpha^2/C_0$. The Hermite derivative identity and trace estimate
  in \citet[Fact~19 and the proof of Lemma~6]{diakonikolas2021proper}
  give $\operatorname{tr}(M)\le p\Ltwonorm[\xmarginal]{q}^2$. It follows that
  \[
    \dim(V)\le
    \frac{C_0p(1+c_0\alpha)^2}{\alpha^2}
    =\bigO{\alpha^{-4}}.
  \]

  \noindent Fix a halfspace $f=h_{w,t}$ and write $w=w_V+w_\perp$ for its
  orthogonal decomposition with respect to $V$. If $w_\perp=0$,
  then $f$ already has its normal in $V$. Otherwise, set
  $\xi=w_\perp/\norm{w_\perp}_2$, and define
  \[
    (A_\xi g)(x)
    =\bE_{G\sim\cN(0,1)}
    \bs{g\br{x-\ip{\xi}{x}\xi+G\xi}}.
  \]
  Under $\xmarginal$, this is conditional expectation given the projection
  onto $\xi^\perp$. Thus, $A_\xi$ is an orthogonal projection in
  $\Ltwo[\xmarginal]$, a contraction in $\Lone[\xmarginal]$, and preserves $\cP_p$.
  Applying \citet[Claim~10]{diakonikolas2021proper} directly gives
  \begin{equation}\label{eq:proper-directional-poincare}
    \Ltwonorm[\xmarginal]{q-A_\xi q}^2
    \le\bE\bs{(\partial_\xi q(X))^2}
    =\xi^\top M\xi
    \le\alpha^2/C_0.
  \end{equation}

  By \Cref{lem:unit-norm-approx-gauss}, taking $C_0$ sufficiently
  large gives $a\in\cP_p$ with $\Ltwonorm[\xmarginal]{a}\le1$ and
  $\Lonenorm[\xmarginal]{f-a}\le\alpha$. Set $Q=a-A_\xi a$. Then
  $\Ltwonorm[\xmarginal]{Q}\le1$, $A_\xi Q=0$, and
  \[
    \Lonenorm[\xmarginal]{(f-A_\xi f)-Q}
    \le\Lonenorm[\xmarginal]{f-a}+\Lonenorm[\xmarginal]{A_\xi(f-a)}
    \le2\alpha.
  \]
  Since $\abs{U}\le1$ and $A_\xi Q=0$, we obtain
  \begin{align*}
    \bE\bs{U(f(X)-A_\xi f(X))}
     & \le2\alpha+\bE\bs{(U-q(X))Q(X)}
    +\bE\bs{(q(X)-A_\xi q(X))Q(X)}     \\
     & \le(2+c_0+C_0^{-1/2})\alpha,
  \end{align*}
  where the last inequality uses
  \Cref{eq:proper-approximate-orthogonality,eq:proper-directional-poincare}.
  Now,
  \[
    A_\xi f(x)
    =\bE_G\bs{\sgn{\ip{w_V}{x}+\norm{w_\perp}_2G+t}}.
  \]
  This is a mixture of halfspaces with unit normals in $V$,
  or constant classifiers if $w_V=0$. Hence one such classifier
  $h^*$ satisfies
  $\bE[U(f(X)-h^*(X))]\le(2+c_0+C_0^{-1/2})\alpha$.
  When $w_\perp=0$, take $h^*=f$.

  For the finite class, use the covering and discretization bounds
  of \citet[Fact~21 and Fact~22 with its proof]{diakonikolas2021proper}, restated in
  \Cref{fact:euclidean-cover,fact:gaussian-halfspace-discretization}.
  Normalize the nonzero centers of an $\alpha/2$-net of the unit
  ball of $V$ to obtain an $\alpha$-net $\cV$ of its unit sphere.
  Set
  \[
    R=\alpha\ceil{K_0\sqrt{\log(1/\alpha)}/\alpha}
  \]
  for a sufficiently large universal constant $K_0$. Define
  \[
    \cH(q)=
    \bc{h_{v,j\alpha}:v\in\cV,\ j\in\bZ,\ \abs{j\alpha}\le R}
    \cup\bc{h_{e_1,R},h_{e_1,-R}}.
  \]
  When $V=\{0\}$, omit the first set. The two additional halfspaces
  approximate the constant classifiers. The discretization bound
  gives an $h\in\cH(q)$ satisfying
  $\bP[h(X)\ne h^*(X)]=\bigO{\alpha}$, including when $h^*$ is
  constant or has threshold outside $[-R,R]$. Thus,
  \[
    \bE\bs{U(f(X)-h(X))}
    \le\bE\bs{U(f(X)-h^*(X))}
    +2\bP[h(X)\ne h^*(X)]
    \le C\alpha.
  \]
  Finally, the covering bound gives
  \[
    \log\abs{\cH(q)}
    =\bigO{(\dim(V)+1)\log(1/\alpha)}
    =\bigO{\alpha^{-4}\log(1/\alpha)}.
  \]
  The construction uses only $q$ and $\alpha$.
\end{proof}

\begin{proof}[Proof of~\Cref{lem:private-l2-polynomial}]
  Let
  \[
    \widehat m=\frac2n\sum_{(X_i,Y_i)\in S_1}
    Y_i\bar\psi_{p,T}(X_i).
  \]
  Each summand has norm at most $T$, so $\widehat m$ has
  replacement sensitivity $4T/n$. The noise scale in
  \Cref{alg:private-proper-gaussian} therefore makes
  $\widehat m+Z$ $(\epsDP,\deltaDP)$-DP
  by standard Gaussian mechanism~\citep{dwork2014algorithmic}.
  Projection and construction of $q$ are post-processing and preserve privacy.

  Write $\ell=\log(8/\beta)$ and
  $L_{\deltaDP}=\log(1.25/\deltaDP)$.
  The centered summands have norm at most $2T$ and second moment
  at most $D_p$. By \Cref{lem:vector-bernstein}, with probability
  at least $1-\beta/4$,
  \[
    \norm{\widehat m-m_T}_2
    \le C_1\br{\sqrt{\frac{D_p\ell}{n}}+\frac{T\ell}{n}}.
  \]
  Gaussian concentration gives, with probability at least
  $1-\beta/4$,  \(\norm{Z}_2\le\sigma\br{\sqrt{D_p}+\sqrt{2\ell}}.
  \)
  Orthonormality and $\abs{U}\le1$ imply $\norm{m_T}_2\le1$.
  Projection onto the unit ball therefore cannot increase distance
  to $m_T$. Using $T=2\sqrt{D_p}$, with probability at least
  $1-\beta/2$ we obtain
  \begin{align*}
    \norm{\widetilde m-m_T}_2
     & \le\norm{\widehat m-m_T}_2+\norm{Z}_2               \\
     & \le C_2\br{
                \sqrt{\frac{D_p\ell}{n}}
                +\frac{\sqrt{D_p}\ell}{n}
                +\frac{(D_p+\sqrt{D_p\ell})\sqrt{L_{\deltaDP}}}
                {\epsDP n}}.
  \end{align*}
  For a sufficiently large universal constant $K$, the condition
  \begin{equation}\label{eq:proper-regression-sample-size}
    n\ge K\br{
      \frac{D_p\ell}{\alpha^2}
      +\frac{D_p\ell\sqrt{L_{\deltaDP}}}
      {\epsDP\alpha}}
  \end{equation}
  makes this at most $c_0c\alpha$. By \Cref{eq:expand-sq-loss},
  \[
    L_s(q)-\inf_{Q\in\cP_p}L_s(Q)
    =\norm{\widetilde m-m_T}_2^2
    \le c_0^2c^2\alpha^2.
  \]
  Finally, projection and orthonormality give
  $\Ltwonorm[\xmarginal]{q}=\norm{\widetilde m}_2\le1$ on every run.
\end{proof}

\begin{proof}[Proof of~\Cref{thm:private-proper-gaussian}]
  Round $n$ up to an even integer. If one observation in $S_1$
  changes, the private polynomial release is followed only by
  post-processing with $S_2$ fixed. If one observation in $S_2$
  changes, the distribution of $q$ is unchanged, and conditional
  on $q$, selection is $\epsDP$-DP. Thus, the
  complete algorithm is
  $(\epsDP,\deltaDP)$-DP.

  Since $p=\ceil{C_0/(c\alpha)^2}$, choosing
  $C_{\mathrm{dim}}\ge2(C_0+1)^2c^{-5}$ ensures
  $d\ge2p^2/(c\alpha)$. Thus, the dimension condition needed in
  \Cref{lem:proper-class-accuracy} follows from the theorem's
  assumption. The first two terms in the sample bound imply
  \Cref{eq:proper-regression-sample-size} and hence the event in
  \Cref{lem:private-l2-polynomial} with probability at least
  $1-\beta/2$. On this event, \Cref{lem:proper-class-accuracy} gives
  \[
    \min_{h\in\cH(q)}\err[\jointdist]{h}
    \le\OPT+(C/2+1)c\alpha.
  \]
  Condition on any such released polynomial and write
  $K=\abs{\cH(q)}$. Applying \Cref{lem:private-finite-validation}
  to the $n/2$ observations in $S_2$ gives, with probability at
  least $1-\beta/2$,
  \[
    \err[\jointdist]{\widehat h}
    \le\min_{h\in\cH(q)}\err[\jointdist]{h}
    +2\sqrt{\frac{\log(8K/\beta)}{n}}
    +\frac{4\log(4K/\beta)}{\epsDP n}.
  \]
  Since $\log K=\bigO{\alpha^{-4}\log(1/\alpha)}$, the last two
  terms in the theorem's sample bound control the dependence on
  $K$. The first two terms also control the remaining confidence
  terms, since $D_p\ge1$. With sufficiently large constants in
  the sample bound, the selection error is at most $c\alpha$.

  A union bound therefore gives, with probability at least
  $1-\beta$,
  \[
    \err[\jointdist]{\widehat h}
    \le\OPT+(C/2+2)c\alpha
    \le\OPT+\alpha,
  \]
  where the last inequality follows by taking $c$ sufficiently
  small. Every member of $\cH(q)$ is a halfspace, so the learner
  is proper.

  Finally, we analyse the running time.
  Computing the clipped Hermite features of the samples in $S_1$ takes $\bigO{npD_p}$ time, and the noisy average and its projection take $\bigO{nD_p}$ more.
  Since $\partial_i h_\kappa = \sqrt{\kappa_i}\, h_{\kappa - e_i}$, the matrix $M = \bE[\nabla q(X)\nabla q(X)^\top]$ can be computed exactly from the coefficients of $q$ in $\bigO{d^2 D_p}$ time, and its eigendecomposition takes $\bigO{d^3}$ time. 
  By~\Cref{lem:proper-finite-class}, $|\cH(q)| = \exp\bigl(\bigO{\alpha^{-4}\log(1/\alpha)}\bigr)$; each member can be listed in \(\operatorname{poly}(d,1/\alpha)\) time and evaluated on \(S_2\) in \(\bigO{nd}\) time. 
  Thus, the exponential mechanism takes \(\bigO{nd|\cH(q)|}\) time overall.
  Since $d \le D_p \le n$ and $1/\alpha \le n$, the total running time is $\operatorname{poly}(n)\cdot\exp\bigl(\bigO{\alpha^{-4}\log(1/\alpha)}\bigr)$.
\end{proof}

\section{Missing Details for Semi-Agnostic Learning}
\label{app:semi-agnostic}
\begin{algorithm}[ht!]
  \caption{DP Semi-Agnostic Halfspace Learner}\label{alg:semi-agnostic}
  \begin{algorithmic}[1]
    \Require Dataset $D=((x_1,y_1),\dots,(x_n,y_n))$,
    privacy parameters $\epsDP,\deltaDP>0$, accuracy parameter $\alpha$, error probability $\beta$.

    \Statex

    \LeftComment{Partition dataset for multiple runs and set parameters.}
    \State Set $K \gets C_K \log(1/\alpha)$ for universal constant $C_K \geq 1$.
    \State Let $D_{val} \subset D$, such that $|D_{val}| = \bigO{\tfrac{\log(K/\beta)}{\alpha^2} + \tfrac{\log(K/\beta)}{\alpha\epsDP}}$. Set $D' \gets D \setminus D_{val}$.
    \State Partition $D'$ into $K$ equal parts $D^1,\dots,D^K$.
    \State Set $\eta_1 \gets C_{\eta}\alpha$, and $\eta_k \gets 2^{k-1} \eta_1$ for $k \in [K]$ and universal constant $C_\eta>1$.
    \Statex

    \LeftComment{Run Algorithm~\ref{alg:semi-agnostic-subroutine} on each partition with appropriate parameters.}
    \For{$k = 1$ to $K$}
    \State Set $h_k \coloneqq (\hat{w}_k,\hat{\tau}_k) \gets \DPSAS_{\epsDP,\deltaDP,\eta_k}(D^k)$.
    \EndFor
    \Statex

    \LeftComment{Run all $h_k$'s on $D_{val}$ and pick the best one using exponential mechanism.}
    \State Let $H \gets (h_1,\dots,h_K)$
    \State Define the score function $\Score(D_{val},h_k) \coloneqq \tfrac{1}{|D_{val}|}\sum\limits_{(x,y) \in D_{val}}\mathbf{1}\{\sgn{\hat{w}_k \cdot x + \hat{\tau}_k} = y \}$.
    \State Set $\Delta \gets 1/|D_{val}|$.
    \State Set $\bar{h} \gets \ExpMech_{\epsDP,H}(\Score,\Delta,D_{val})$.
    \Statex

    \State \textbf{Return} $\bar{h}$.

  \end{algorithmic}
\end{algorithm}

\subsection{Privacy}

We first prove the privacy guarantees of Algorithm~\ref{alg:semi-agnostic}.
\begin{proposition}\label{prop:semi-agnostic-privacy}
  Algorithm~\ref{alg:semi-agnostic} satisfies $(\bigO{\epsDP},\bigO{\deltaDP})$-DP.
\end{proposition}
\begin{proof}
  We show that all different steps of Algorithm~\ref{alg:semi-agnostic} that involve referencing the data using Algorithm~\ref{alg:semi-agnostic-subroutine} and the exponential mechanism satisfy DP, and then we apply parallel composition. However, for that, we first need to prove the privacy guarantees of Algorithm~\ref{alg:semi-agnostic-subroutine} by combination of using privacy guarantees of the standard DP mechanisms and parallel composition.

  In Algorithm~\ref{alg:semi-agnostic-subroutine}, when computing the threshold $\hat{\tau}$, the sensitivity of $\mathbb{E}_{(x,y) \sim D}[y]$ (the average of all the labels) has sensitivity $2/m_0$ because the labels can be either $-1$ or $1$. Therefore, adding Laplace noise scaled to $2/\epsDP m_0$ satisfies $\epsDP$-DP.

  For computing the initial weight vector $\hat{w}_0$, we compute the noisy average of $xy$ over the entire dataset $D_{init}$. Since we clip all the data points to within $L$ in $\ell_2$ norm, and $y$ is either $-1$ or $1$, the $\ell_2$ sensitivity of this average is $2L/m_0$. Thus, adding Gaussian noise to this average scaled to $2L\sqrt{2\log(2/\deltaDP)}/\epsDP m_0$ satisfies $(\epsDP,\deltaDP)$-DP.

  Next, in each iteration of the gradient descent procedure, we add Gaussian noise to the average gradient vector. From Claim~\ref{clm:noisy-grad-def}, we see that the average gradient is of the form
  $$\mathbb{E}_{(x,y) \sim D}\left[-y \mathbf{1}\{|w\cdot x -b_t| \leq a_t\} \nabla_w\tfrac{w\cdot x}{\|w\|}\right].$$
  Therefore, the sensitivity of this average gradient is at most
  $$\frac{2\left\|\nabla_w\tfrac{w\cdot x}{\|w\|}\right\|}{m_0} \leq \frac{2\|x\|}{m_0} \leq \frac{2L}{m_0}.$$
  This implies that adding Gaussian noise to the average scaled to $2L\sqrt{2\log(2/\deltaDP)}/\epsDP m_0$ satisfies $(\epsDP,\deltaDP)$-DP in each iteration. However, in each iteration, a disjoint dataset gets used, so no data point gets used twice for any DP operation. Hence, by parallel composition, we get $(\epsDP,\deltaDP)$-DP for this part, as well.

  Finally, applying parallel composition over the three steps again, we obtain $(\epsDP,\deltaDP)$-DP for Algorithm~\ref{alg:semi-agnostic-subroutine}.

  The guarantees of the exponential mechanism in Algorithm~\ref{alg:semi-agnostic} imply $\epsDP$-DP because the sensitivity of the mean of $0$-$1$ random variables is $1/|D_{val}|$. Since all the DP steps in Algorithm~\ref{alg:semi-agnostic} are taken on disjoint and separate datasets, we obtain the required DP guarantee using parallel composition.
\end{proof}

\subsection{Relevant Assumptions and Regularity Conditions}\label{sec:semi-agnostic-reg-cond}

For simplicity, we assume that the ambient dimension $d$ is large enough. This assumption can be removed by using the exact expressions for concentration inequalities for high-dimensional Gaussians.
\begin{assumption}[Assumption on ambient dimension]\label{ass:det-reg}
  We assume the following bound on the ambient dimension in terms of the error probability $\beta \in (0,1)$ and number of samples $n \in \mathbb{N}$: $d \geq 16\log(n/\beta)$.
\end{assumption}

We do this for ease of notation, but we can make our calculations exact with a few extra lower order terms by not making this assumption.

Next, we state the Laurent-Massart inequality for concentration of high-dimensional Gaussians.
\begin{lemma}[Laurent-Massart inequality]\label{lem:laurent-massart}
  Let $X \sim \mathcal{N}(0,\mathbb{I})$ be $d$-dimensional, and $\Sigma = A^\top A$ for $A \in \mathbb{R}^{d \times d}$. Then the following holds for all $t > 0$.
  \begin{align*}
    \Pr[\|AX\|_2^2 > \mathrm{tr}(\Sigma) + 2\sqrt{\mathrm{tr}(\Sigma^2)t} + 2t\|\Sigma\|_2] & \leq e^{-t} \\
    \Pr[\|AX\|_2^2 < \mathrm{tr}(\Sigma) - 2\sqrt{\mathrm{tr}(\Sigma^2)t}] \leq e^{-t}
  \end{align*}
\end{lemma}

Using the above, we have the following deterministic regularity condition that we would assume throughout the proof.
\begin{lemma}[Concentration of high-dimensional Gaussians]\label{lem:gauss-conc-high-d}
  Let $X_1,\dots,X_n$ be i.i.d.~samples from a $d$-dimensional Gaussian $\mathcal{N}(0,\mathbb{I})$. Then for $\beta \in (0,1)$, the following holds with probability at least $1-\beta$.
  $$\Pr\left[\forall~i \in [n]: \|X_i\| \in [\sqrt{d}/2, 2\sqrt{d}]\right] \geq 1-\beta$$
\end{lemma}
\begin{proof}
  The proof follows by simply applying Assumption~\ref{ass:det-reg}, Lemma~\ref{lem:laurent-massart}, and the union bound over all $n$ samples.
\end{proof}

In Algorithm~\ref{alg:semi-agnostic-subroutine}, we set the clipping parameter $L=2\sqrt{d}$ to ensure that all data points lie completely within the $\ell_2$ ball of radius $L$ around the origin, and no point gets actually clipped with high probability.

\subsection{Missing Proofs from Section~\ref{sec:semi-agnostic-proper-gaussian}}

\begin{proof}[Proof of Claim~\ref{clm:noisy-grad-def}]
  The proof of this claim is the same as that of Claim~C.2 from~\citep{DiakonikolasKTN22}, except that the noisy gradient term has Gaussian noise added in the subspace orthogonal to $w$ (which is a known DP quantity) because the weight vectors get ultimately projected onto the unit ball and the GD algorithm will not move in the direction of $w$. Therefore, by Pythagoras' theorem, $w - \lambda\nabla_w \tilde{\mathcal{L}}_t(w)$ term will still have norm at least $1$ as the gradient and the noise belong to the same subspace that is orthogonal to $w$.
\end{proof}

\begin{proof}[Proof of Claim~\ref{clm:semi-agnostic-initialization}]
  In our algorithm $\tilde{w}_0 = w_0 + z_0$, where $z_0 \sim \mathcal{N}(0,\sigma^2\mathbb{I})$, $\sigma = \tfrac{4\sqrt{2d\log(2/\deltaDP)}}{\epsDP m_0}$, and $w_0$ is as defined by~\cite{DiakonikolasKTN22} in Algorithm~2. We know from the proof of Theorem~1.2 in~\citep{DiakonikolasKTN22} that $w^* \cdot w_0 > \omega(\eta)$. Therefore, it is sufficient to bound $w^*\cdot z_0$ because $w^* \cdot \tilde{w}_0 >0 \implies w^* \cdot \hat{w}_0 > 0$. Note that for any fixed unit vector $w^*$, $w^*\cdot z_0 \sim \mathcal{N}(0,\sigma^2)$. Applying concentration inequality for 1-D Gaussians, we get that with probability at least $1-\bigO{\beta}$, $|w^* \cdot z_0| \leq \bigO{\tfrac{\sqrt{d\log(1/\deltaDP)\log(1/\beta)}}{\epsDP m_0}} < \bigO{\eta}$ by plugging in our bound on the sample complexity.
\end{proof}

\end{document}